\documentclass[10pt,twocolumn,letterpaper]{article}

\usepackage{cvpr}              %

\definecolor{cvprblue}{rgb}{0.21,0.49,0.74}
\usepackage[pagebackref,breaklinks,colorlinks,allcolors=cvprblue]{hyperref}
\usepackage{cuted}
\usepackage{caption}
\usepackage{xcolor}
\usepackage{adjustbox}
\usepackage{array}
\usepackage{tabularx, booktabs}

\def\confName{CVPR}
\def\confYear{2027}

\title{Code2RSI: Towards Recursive Self-Improvement for \\ Real-to-Sim Code World Models}

\author{
Jingyu Hu$^{1*}$ \quad
Haoxuan Cao$^{2*}$ \quad
Zeyu Zhang$^{2*\dag}$ \quad
Mingju Gao$^{2}$ \quad
Hao Tang$^{2\ddag}$ \\
$^{1}$University of Bristol \quad
$^{2}$School of Computer Science, Peking University\\
\small $^*$Equal contribution.
$^\dag$Project lead.
$^\ddag$Corresponding author: bjdxtanghao@gmail.com.
}

\begin{document}
\maketitle

\begin{strip}
\vspace{-2em}
\centering
\noindent
\includegraphics[width=\linewidth,trim=0 1bp 0 0,clip]{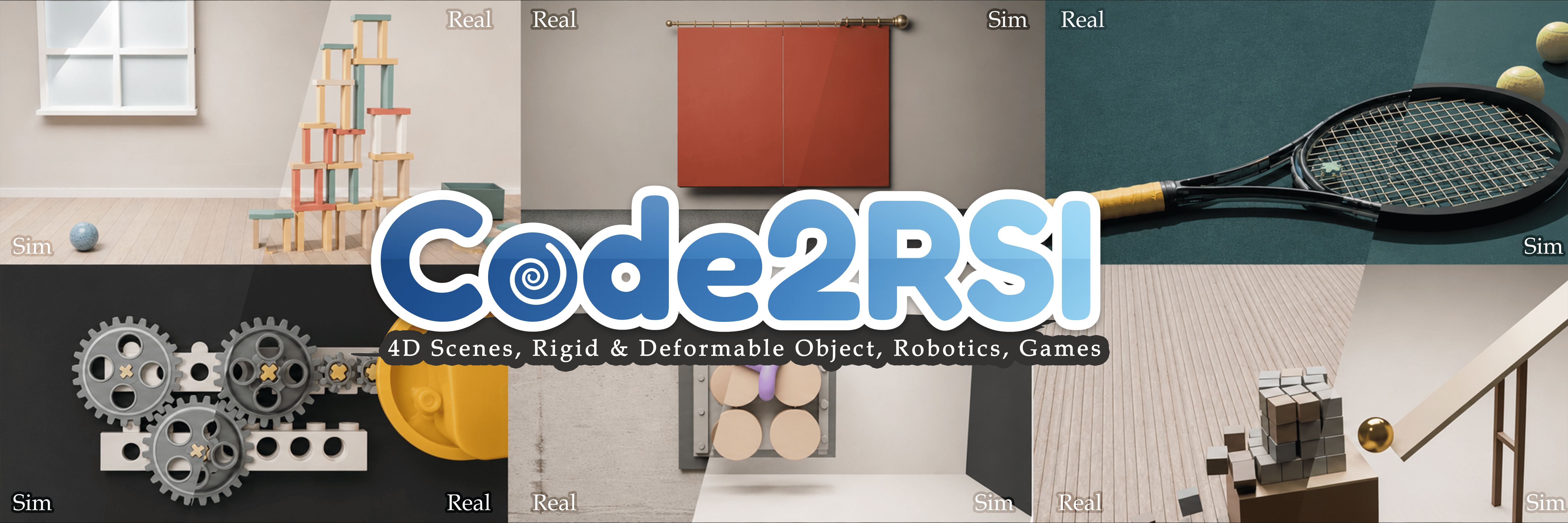}
\captionsetup{hypcap=false}
\captionof{figure}{Examples of Real-to-sim Code-based World Construction. Each panel contrasts a world before and after construction with Code2RSI, covering rigid dynamics, deformable objects, articulated mechanisms and robotic interaction.}
\captionsetup{hypcap=true}
\label{fig:teaser}
\end{strip}

\begin{abstract}
Generative models have shown promise in constructing interactive world.
However, these worlds can appear visually plausible and exhibit desired outcomes without faithfully modeling the underlying physical interactions that produce them. 
This gap between visual plausibility and satisfaction of task requirements remains a challenge for reliable Real-to-Sim.
We present Code2RSI, an agentic framework that structures world construction around task-specific agent harnesses, each connecting world programs with domain-specific tools, observations and feedback. 
Code2RSI enables recursive self-improvement by distinguishing immediate repairs to individual worlds from persistent updates to reusable construction skills and harness routines.
We evaluate the framework on dynamic scene reconstruction and robotic interaction.
Code2RSI improves reconstruction quality and task performance in these settings. Additional experiments on text-conditioned scene generation and task-driven game construction demonstrate its applicability beyond Real-to-Sim.
These findings highlight the potential of execution-driven recursive self-improvement for reliable world construction.
\href{https://github.com/AIGeeksGroup/Code2RSI}{Code} \&
\href{https://aigeeksgroup.github.io/Code2RSI}{Website}.
\end{abstract}

\section{Introduction}
\label{sec:intro}

\begin{figure*}[t]
    \centering
    \includegraphics[width=0.95\linewidth]{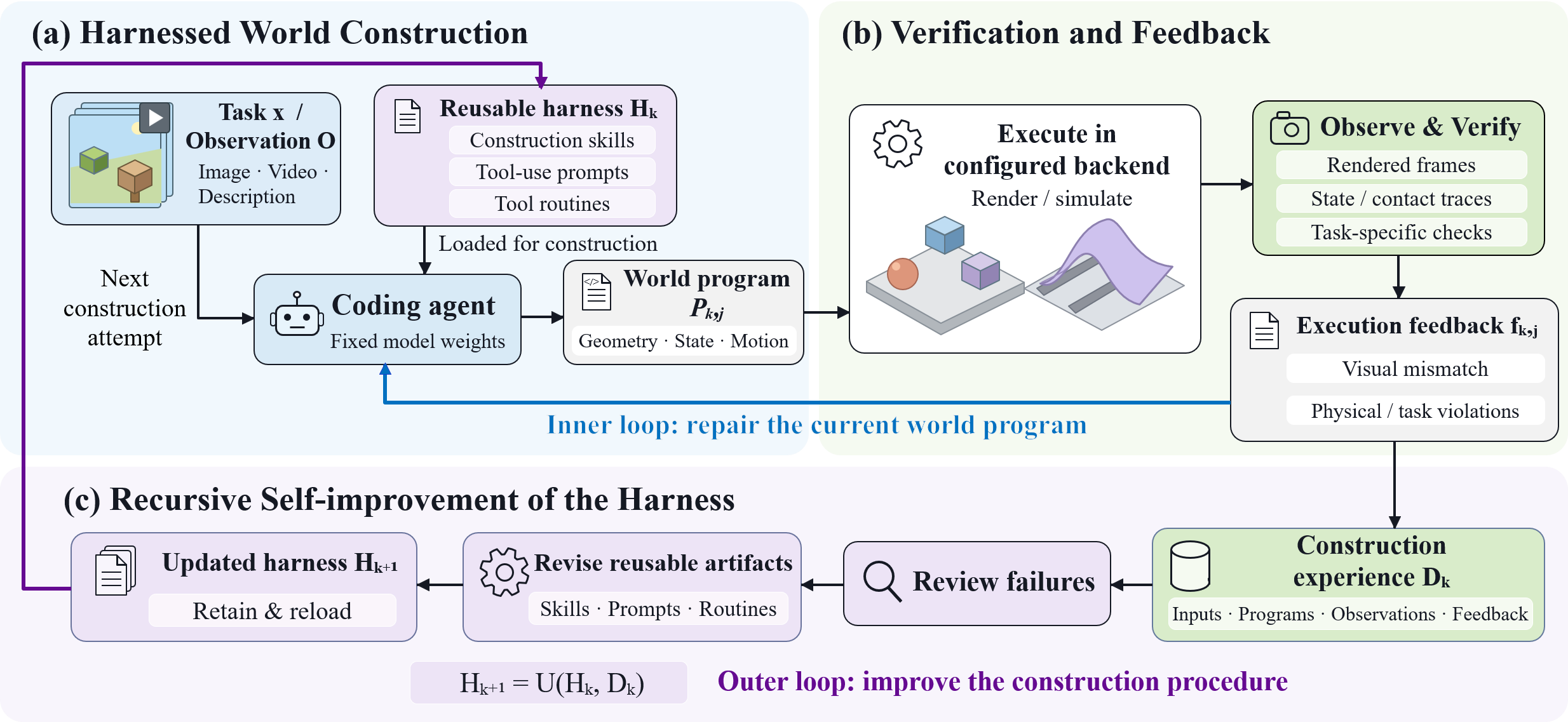}
    \caption{Overview of Code2RSI. A task-specific harness connects the coding agent to execution, observation, and verification. The inner loop repairs the current world program, while the outer loop uses construction records to update reusable skills and harness routines.}
    \label{fig:workflow}
\end{figure*}

Real-to-Sim aims to construct executable simulations of real-world scenes and behaviorss~\cite{ye20263d,wong2025survey,pfaff2025scalable}. These world-construction tasks require representations of scene structure, object behavior, and dynamic interactions. 
Advances in coding agents create opportunities to generate world programs from task descriptions and reference observations~\cite{chen2026agenticreal2sim,xu2026worldmirror,tang2026video2world, wang2026scenecode}. 
Despite progress in generating executable world programs, successful execution does not ensure that a generated world preserves the requested content, supports task completion, or satisfies relevant physical constraints~\cite{peng2026exeworldbench,nishizawa2026yubi}.
For Real-to-Sim, reliability concerns both the fidelity of reconstructed scenes and dynamics, and the validity of interaction behaviors executed in simulation. This raises the question: \emph{How can coding agents construct reliable executable scenes and interaction behaviors for Real-to-Sim Tasks?}

Recent work ~\cite{shen2026fourdcodebench,shao2026physcodegym,tang2026video2world} has started examining these limitations in dynamic scene reconstruction and robotic interaction.
Specifically, 4DCodeBench~\cite{shen2026fourdcodebench} finds that strong static reconstruction cannot be translated into reliable reconstruction of complex dynamics.
Code2Worlds~\cite{zhang2026code2worlds,zhang2026code2worlds2} identifies missing physical properties and scripted motion as limitations of visually plausible generated worlds.
In dynamic scene reconstruction, visual similarity alone does not establish that scene geometry and motion have been faithfully recovered. In robotic interaction, task completion alone does not establish that the intended physical interactions occurred. These limitations motivate separate assessments of visual, task, and physical reliability.

Assessing these requirements requires evidence beyond the final visual output. For dynamic scene reconstruction, rendered frames and motion traces can show complementary errors in appearance and dynamics. For robotic interaction, task states and contact records can help determine if successful outcomes were produced through valid interactions.
However, execution environments expose this information through different interfaces, and checks designed for one task may not apply to another.
An agent therefore needs task-specific tools to execute world programs, collect the relevant observations, and check them against the task requirements.
To improve reliability, the problems identified by these checks must also be returned as feedback that can guide program revision~\cite{xu2026worldmirror,shao2026physcodegym,zhang2026code2worlds2}.
These requirements motivate task-specific agent harnesses that connect execution tools, reliability checks, and feedback within a shared workflow for world program construction, execution, verification and revision.

We propose \textbf{Code2RSI}, an agentic framework for reliable code-based world construction, with Real-to-Sim as its primary application. We investigate two complementary settings: dynamic scene reconstruction, where executable scene programs recover geometry and motion from reference videos, and robotic interaction, where control programs produce task-directed behaviors in simulated scenes. Task-specific harnesses connect coding agents to the execution tools, observations and reliability checks required by each setting.
Code2RSI builds on execution-guided program revision~\cite{xu2026worldmirror,shao2026physcodegym} and harness-based recursive self-improvement (RSI)~\cite{feng2026enactopt,lee2026emho,xia2026rrsi}. The inner loop uses execution feedback to repair the current world program, while the outer loop retains updates to reusable construction skills and harness routines for subsequent construction attempts. We further examine the effectiveness of the framework beyond Real-to-Sim through text-conditioned scene generation and task-driven game construction.

With GPT-6-Astra, Code2RSI raises Dynamic IoU from 0.355 to 0.584 and trajectory DTW from 0.734 to 0.901 in dynamic scene reconstruction. 
Additional experiments on text-conditioned scene generation and task-driven game construction in Minecraft show its applicability beyond Real-to-Sim. Our analysis also shows that improvements depend on what the verification procedure measures: higher aggregate scores do not guarantee better performance on physical criterion.

Overall, our contributions can be concluded as follows.
\begin{itemize}

\item \textbf{A code-based construction framework for Real-to-Sim.} Code2RSI supports dynamic scene reconstruction and robotic interaction through executable programs, and connects coding agents to execution tools and reliability checks through task-specific agent harnesses.
\item \textbf{Execution-Grounded Recursive Self-Improvement}. We introduce the agent loop mechanism where an inner loop revises the current world program against execution feedback, and an outer loop distills the revision records into retained updates of modeling harness routines. This closed-loop system iteratively turns failures in one construction into guidance for improving reliability without updating model weights.

\item \textbf{Diverse application scenarios.} Code2RSI demonstrates reliability improvements in dynamic scene reconstruction and robotic interaction. Its extension to text-conditioned scene generation and task-driven game construction also shows promising results beyond Real-to-Sim. These findings suggest potential for broader applications in reliable world construction.

\end{itemize}

\section{Related Work}
\label{sec:related-work}

\subsection{Code-Based Real-to-Sim World Construction}

Executable world programs support both reconstruction from observations and construction from task descriptions~\cite{xu2026worldmirror,chen2026robosimstudio,ren2026real2gym,tang2026video2world}.
For real-to-sim reconstruction, VIGA~\cite{yin2026viga} and LEGO-Anything~\cite{li2026legoanything} recover editable scenes through code-render-inspect loops, while 4DCodeBench~\cite{shen2026fourdcodebench} evaluates whether agents can encode video dynamics as executable graphics programs.
Agentic Real2Sim~\cite{chen2026agenticreal2sim} connects reconstruction to robot interaction by assembling observed episodes into runnable physical simulations.
For task-driven generation, SceneSmith~\cite{pfaff2026scenesmith} constructs simulation-ready indoor scenes, and Code2Games~\cite{wu2026code2games} generates worlds together with executable gameplay.
Code World Model~\cite{chen2026codeworld} further connects program execution with visual synthesis by maintaining world state in code and conditioning video generation on proxy renders.

\subsection{Execution-Guided Verification and Revisions}

The execution environment determines which evidence can guide program repair.
Code2Games~\cite{wu2026code2games} localizes scene and gameplay errors using engine diagnostics, gameplay assertions, and interaction traces.
For simulated interactions, Code2Worlds-2~\cite{zhang2026code2worlds2} uses simulator and visual critics to revise dynamics programs and physical settings, while PhysAgent~\cite{li2026physagent} couples simulation with stage-specific verification and targeted repair.
PhysCodeGym~\cite{shao2026physcodegym} evaluates physical consistency through program inspection, scene measurements, and counterfactual interventions.
Code2RSI organizes these forms of evidence through task-specific harnesses: rendered observations, task states, and physical traces guide world program revision within a shared construction workflow.

\subsection{Recursive Self-Improvement of Agents and Harness Engineering}

Recursive self-improvement extends feedback from revising individual outputs to retaining changes in the procedures governing later executions~\cite{fan2026videoharness,feng2026enactopt,lee2026emho}.
RRSI~\cite{xia2026rrsi} regularizes the proposal and selection of harness edits around a frozen model; Video-RSI~\cite{luo2026videorsi} grounds reusable harness revisions in investigations of video-understanding failures and selects them according to accuracy and visual cost.
In robotics, EmbodiRSI~\cite{lang2026embodirsi} instead uses simulation feedback to adapt corrective experience collection and update manipulation policies.
Code2RSI shares the selection rule of these methods and differs in what is updated and what selects it. The update edits the modeling guidance and the skills that include executable physical checks, and a candidate is retained only when the worlds it constructs verify better on the task.

\section{Framework}
\label{sec:method}

Code2RSI organizes world construction around task requirements, execution and feedback. We first formulate the construction problem and define three aspects of reliability, then describe the harness and recursive self-improvement procedures used to revise programs and retain reusable updates.
Code2RSI runs as two nested loops. The inner loop revises the current world program under the current harness. The outer loop is the recursive self-improvement step which uses the records of the inner loop to update the harness.

\subsection{Problem Formulation}
\label{sec:scope}

A task $x$ specifies the scene or interaction to be constructed. In dynamic scene reconstruction, it includes reference observations, such as frames or videos, and reconstruction requirements. In robotic interaction, it specifies a task objective and provides access to a simulated scene.
Given $x$, our coding agent $A$ constructs a world program $P$ that represents scene structure and dynamics or controls interactions within the scene.
The program is executed by a backend $E$, such as a renderer or physical simulator. We denote the execution outcomes such as rendered frames, state traces, or action outcomes, as the observation $O = E(P)$. 
Feedback on the world is obtained through verification. A verification procedure $\phi(x, P, O)$ compares the observation with the task requirement, the reference observations, or the intended state held in $P$, and returns a measurement or a verdict. 
Part of this feedback is returned to $A$ and leads to the next round revision,
and the rest is computed after construction and reported as results. 
The reliability of a constructed world is a property of $(P, E)$ with respect to $x$ judged based on $O$. 
We characterize reliability in Real2Sim via three aspects. \emph{Visual reliability} is about the agreement between the reconstructed scene and reference observations in appearance and spatial layout.
\emph{Task reliability} refers to successful program execution and completion of the specified manipulation objective. \emph{Physical reliability} concerns whether reconstructed worlds satisfy relevant physical constraints during execution.

\subsection{World Construction Harness}
\label{sec:method-harness}

The reliability requirements above determine what the agent needs to observe before revising a world program. 
Appearance errors can be inspected in rendered frames, an unsuccessful action or an invalid contact may require task states or interaction traces. 
A task-specific harness supplies the tools, execution access, and checks needed to obtain this evidence. We denote its harness configuration by $H_k$, where $k$ counts retained updates to modeling skills and tool-use routines.
The agent constructs the world program $P_{k,j}$ using the representations and tools supported by the chosen environment, where $j$ indexes world revisions under $H_k$.

The harness then executes it through a compatible backend and collects observation $O_{k,j}=E(P_{k,j})$. 
For dynamic scene reconstruction, rendered frames and exported trajectories provide evidence of appearance and motion. For robotic interaction, simulator states and interaction traces provide evidence of task outcomes and physical constraints.
Table~\ref{tab:domain-setup} in the appendix summarizes the representations and execution interfaces for these settings and the applications beyond Real2Sim.
To turn these observations into guidance for revision, the verification procedure compares the program and its execution with the task requirements:
\begin{equation}
 f_{k,j}=\phi(x,P_{k,j},O_{k,j}).
 \label{eq:feedback}
\end{equation}
Given this feedback, the inner loop revises the program under the current harness $H_k$. The agent modifies scene geometry, dynamics, or control logic according to the task:
\begin{equation}
 P_{k,j+1}=A(x,P_{k,j},f_{k,j},H_k).
 \label{eq:program-update}
\end{equation}
The harness executes the revised program and checks its observations again, allowing the next revision to respond to the effects of the previous change.

\subsection{Recursive Self-Improvement}
\label{sec:rsi}

Execution feedback can improve a particular world program, but the experience behind a repair is not automatically available to subsequent construction attempts. If only the world program changes, the agent will still rely on the same modeling instructions and tool routines when constructing the next program. 
This motivates retaining reusable changes to the construction procedure itself. 
Inspired by ~\cite{feng2026enactopt,lee2026emho,xia2026rrsi}, Code2RSI uses feedback both to revise the current program and to update the configuration that guides later attempts. 
The outer loop makes the information available beyond the current world. RSI uses selected construction records to revise $H_k$. Let $\mathcal{D}_k$ contain records $(x,P_{k,j},O_{k,j},f_{k,j})$,
the update procedure $U$ uses these records to revise reusable modeling guidance and harness routines:
\begin{equation}
 H_{k+1}=U(H_k,\mathcal{D}_k).
 \label{eq:harness-update}
\end{equation}
Here $U$ has three steps. Each outer round first collects $\mathcal{D}_k$ from the inner loop on the tasks and the current guidance and skills. The agent then diagnoses the recorded failures and proposes one candidate that edits only these two parts, such as a modeling instruction or an executable check for later programs. The model, backend and verification procedure stay fixed. The candidate is run on the same tasks with the same inner budget and becomes $H_{k+1}$ only when its mean score exceeds that of $H_k$. Otherwise $H_k$ is restored and the rejected candidate enters the next diagnosis. 
The diagnosis step also writes physical checks, executes them on the constructed worlds, and reports any violations to the next proposal round.

\begin{table*}[t!]
  \centering
  \footnotesize
  \setlength{\tabcolsep}{2pt}
  \renewcommand{\arraystretch}{1.18}
  \caption{Performance Comparison of Code2RSI and the Fixed Harness Across Three Real2Sim Domains. We highlight the \textbf{best} values within each backbone and metric. All metrics are higher the better.}
  \label{tab:real2sim-three-domain-results}
  \resizebox{0.9\textwidth}{!}{%
  \begin{tabular}{*{13}{c}}
    \toprule
    & & \multicolumn{4}{c}{Dynamic Scene} & \multicolumn{3}{c}{Deformable Object} & \multicolumn{4}{c}{Robotic Interaction} \\
    \cmidrule(lr){3-6} \cmidrule(lr){7-9} \cmidrule(lr){10-13}
    Model & Method & \mbox{DINOv3} & \mbox{IoU} & \mbox{Scene 3D} & \mbox{DTW} & \mbox{DINOv3} & \mbox{IoU} & \mbox{SS} & \mbox{DINOv3} & \mbox{IoU} & \mbox{TC} & \mbox{PS} \\
    \midrule
    & Harness (w/o RSI) & 0.898 & 0.355 & \textbf{0.880} & 0.734 & 0.466 & 0.143 & 2.4 & \textbf{0.963} & 0.508 & 0.714 & 0.814 \\
    \raisebox{\dimexpr-.5\height+.5ex\relax}[0pt][0pt]{\shortstack{GPT-6-\\Astra}} & Code2RSI ($r{=}1$) & 0.884 & 0.459 & 0.836 & \textbf{0.908} & 0.678 & 0.123 & 4.3 & 0.962 & \textbf{0.554} & \textbf{0.857} & \textbf{0.857} \\
    & Code2RSI ($r{=}2$) & \textbf{0.907} & \textbf{0.584} & 0.868 & 0.901 & \textbf{0.696} & \textbf{0.151} & \textbf{4.7} & 0.961 & 0.476 & 0.714 & 0.856 \\
    \midrule
    & Harness (w/o RSI) & 0.450 & 0.000 & 0.449 & 0.177 & 0.314 & 0.007 & 1.2 & 0.942 & 0.399 & 0.429 & 0.756 \\
    \raisebox{\dimexpr-.5\height+.5ex\relax}[0pt][0pt]{\shortstack{GPT-6\\Luna}} & Code2RSI ($r{=}1$) & 0.436 & 0.000 & 0.457 & 0.263 & \textbf{0.358} & 0.012 & \textbf{1.4} & \textbf{0.948} & \textbf{0.469} & 0.429 & \textbf{0.808} \\
    & Code2RSI ($r{=}2$) & \textbf{0.871} & \textbf{0.071} & \textbf{0.836} & \textbf{0.589} & 0.309 & \textbf{0.100} & 1.2 & 0.820 & 0.358 & \textbf{0.571} & 0.619 \\
    \midrule
    & Harness (w/o RSI) & 0.428 & 0.208 & 0.427 & 0.408 & 0.298 & 0.003 & 1.1 & \textbf{0.942} & 0.413 & 0.571 & 0.619 \\
    \raisebox{\dimexpr-.5\height+.5ex\relax}[0pt][0pt]{\shortstack{DeepSeek-\\V4.1-Flash}} & Code2RSI ($r{=}1$) & \textbf{0.841} & \textbf{0.360} & \textbf{0.860} & 0.752 & 0.486 & 0.016 & 3.0 & 0.822 & 0.406 & 0.571 & 0.700 \\
    & Code2RSI ($r{=}2$) & 0.828 & 0.142 & 0.854 & \textbf{0.829} & \textbf{0.597} & \textbf{0.026} & \textbf{3.5} & 0.940 & \textbf{0.433} & \textbf{0.714} & \textbf{0.782} \\
    \midrule
    & Harness (w/o RSI) & 0.429 & 0.000 & 0.129 & 0.000 & 0.239 & 0.003 & 1.0 & 0.547 & 0.264 & 0.286 & 0.482 \\
    \raisebox{\dimexpr-.5\height+.5ex\relax}[0pt][0pt]{\shortstack{GLM-5.3\\Flash}} & Code2RSI ($r{=}1$) & \textbf{0.846} & 0.001 & 0.369 & \textbf{0.467} & \textbf{0.366} & \textbf{0.037} & \textbf{1.9} & 0.675 & 0.264 & 0.286 & 0.471 \\
    & Code2RSI ($r{=}2$) & 0.414 & \textbf{0.136} & \textbf{0.419} & 0.430 & 0.353 & 0.027 & \textbf{1.9} & \textbf{0.948} & \textbf{0.445} & \textbf{0.571} & \textbf{0.668} \\
    \midrule
    & Harness (w/o RSI) & 0.254 & 0.000 & 0.404 & 0.000 & \textbf{0.425} & \textbf{0.042} & 2.6 & 0.663 & 0.329 & 0.286 & 0.493 \\
    \raisebox{\dimexpr-.5\height+.5ex\relax}[0pt][0pt]{\shortstack{Qwen3.8\\27B}} & Code2RSI ($r{=}1$) & 0.739 & \textbf{0.217} & 0.718 & 0.579 & 0.422 & 0.019 & \textbf{2.7} & 0.676 & 0.310 & \textbf{0.571} & 0.571 \\
    & Code2RSI ($r{=}2$) & \textbf{0.868} & 0.033 & \textbf{0.835} & \textbf{0.637} & 0.360 & 0.010 & 2.5 & \textbf{0.683} & \textbf{0.361} & \textbf{0.571} & \textbf{0.713} \\
    \bottomrule
  \end{tabular}%
  }

\end{table*}

\begin{figure*}[t]
  \centering
  \includegraphics[width=\textwidth]{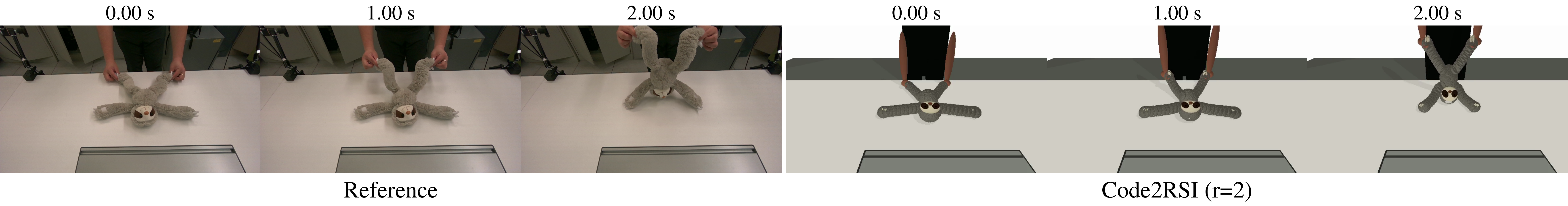}
  \caption{Keyframe Comparison of the Deformable Object Reconstruction with Reference Frames and Frames Generated by Code2RSI.}
  \label{fig:phystwin-sloth-keyframes}
\end{figure*}

\begin{figure*}[t!]
  \centering
  \includegraphics[width=\textwidth]{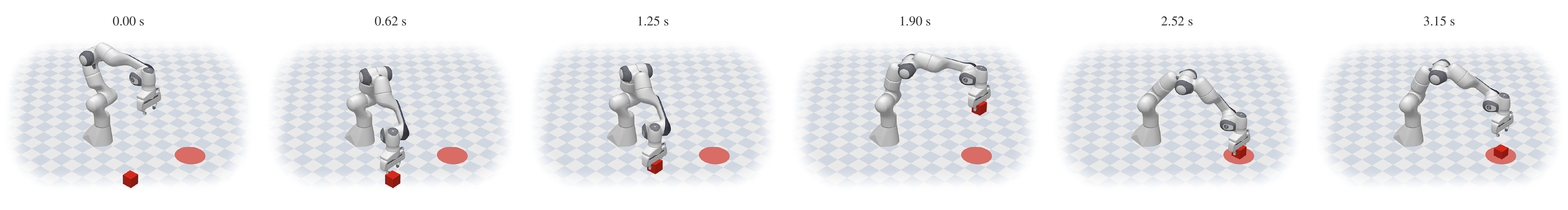}
  \caption{Keyframes of Code2RSI's Generated Policy on the Task of `Pick up the cube, place it at the target marker and release it.' }
  \label{fig:robot-pick-keyframes}
\end{figure*}

\begin{figure*}[t!]
  \centering
  \includegraphics[width=\textwidth]{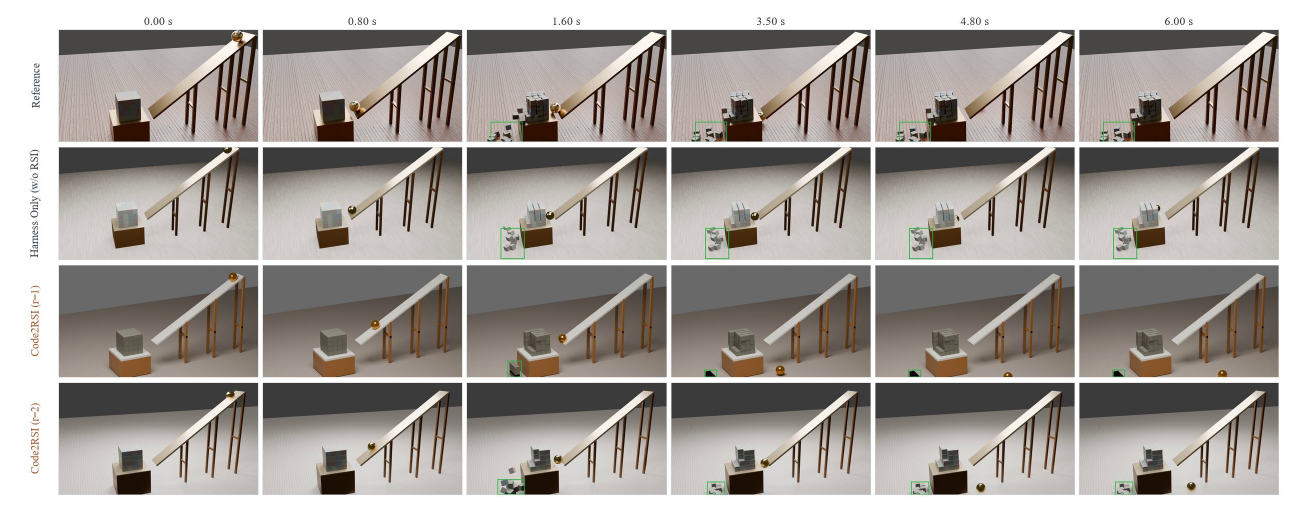}
  \caption{Visual Comparison of Reference Frames, Harness Only (w/o RSI), and Code2RSI ($r=1,2$) Using GPT-6-Astra. Green boxes highlight fallen blocks.}
  \label{fig:rsi-refinement-keyframes}
\end{figure*}

\section{Experiments}
\label{sec:experiments}

We evaluate reconstruction fidelity and interaction reliability in Real2Sim tasks (Sec.~\ref{sec:exp-real2sim}), perform ablation studies (Sec.~\ref{sec_ablation}), and explore broader applications beyond Real2Sim (Sec.~\ref{sec:exp-beyond-real2sim}).

\subsection{Experimental Setup}
\label{sec:exp-setup}

\textbf{Tasks and metrics.} 
We evaluate Code2RSI on dynamic scene, deformable object and robotic interaction reconstruction for Real2Sim, with extensions to text-conditioned scene generation and task-driven game construction. These settings referred 4DCodeBench~\cite{shen2026fourdcodebench}, PhysTwin~\cite{jiang2025phystwin}, PyBullet manipulation tasks~\cite{coumans2016pybullet}, PhyGenBench captions~\cite{meng2024towards}, and Minecraft tasks~\cite{fan2022minedojo} respectively.
We evaluate reliability along three aspects.
\textbf{Visual reliability} is assessed through appearance and geometric agreement: DINOv3~\cite{simeoni2025dinov3} feature similarity measures the appearance of the reconstructed body, and Scene 3D measures its geometry. Agreement with the reference
dynamics is captured by Dynamic IoU, which measures how well the deformation over the clip is recovered, and by trajectory DTW, reported as one minus the normalized DTW distance.
\textbf{Task reliability} is measured by task completion (TC) and progress score (PS). \textbf{Physical reliability} is examined separately through diagnostic checks on the reconstructed geometry and trajectories. For deformable-object reconstruction, we additionally report SS, a vision-language verifier score for visual and physical consistency. Implementation details are provided in Appendix~\ref{app:real2sim-metrics}.

\textbf{Methods.} We evaluate Code2RSI across successive RSI rounds and compare it with Harness (w/o RSI), a variant that retains the inner-loop program repair but fixes the harness at its initial configuration. The Real2Sim experiments use GPT-6-Astra~\cite{openai2026gpt6}, GPT-6 Luna~\cite{openai2026gpt6}, DeepSeek-V4.1-Flash~\cite{deepseek2026v4flash}, GLM-5.3-Flash~\cite{zai2026glm53flash}, and Qwen3.8-27B~\cite{qwen2026qwen38} as backbone models. Comparisons within each setting use the same backbone and inner-loop budget.

\subsection{Application on Real2Sim}
\label{sec:exp-real2sim}

Code2RSI is evaluated on three real2sim settings that cover rigid dynamics, deformable objects and robotic interaction.

\subsubsection{Dynamic Scene Reconstruction}
\label{sec:exp-dynamic-scene}

Given a reference video, the task of dynamic scene reconstruction is to construct an executable scene program for reproduction.

We first assess if RSI improves the recovery of scene appearance and dynamics across different coding backbones. 
After two rounds of RSI, all five models outperform the fixed harness on trajectory DTW, and four of the five on DINOv3. GLM-5.3-Flash peaks on DINOv3 after the first round at 0.846 and falls to 0.414 in the second, below its fixed-harness value of 0.429.
The weaker models gain the most in appearance: GPT-6 Luna rises from 0.45 to 0.871 and Qwen3.8-27B from 0.254 to 0.868.
The gap between models is much larger in Dynamic IoU. GPT-6-Astra improves from 0.355 to 0.584, while GPT-6 Luna and Qwen3.8-27B reach only 0.071 and 0.033.
These results indicate that a high appearance score does not mean that the scene dynamics have been recovered. 
Different metrics can also peak at different rounds. An update can be retained for its appearance gains while the recovered motion remains poor. For example, DeepSeek-V4.1-Flash achieves its best Dynamic IoU after the first RSI round, whereas its trajectory DTW continues to improve in the second. 

To illustrate these gains qualitatively, Figure~\ref{fig:rsi-refinement-keyframes} compares keyframes from the reference sequence with reconstructions across different method variants on the falling-block scene.
The final arrangement shows how the blocks separate and settle after impact. 
In the earlier reconstructions, the displaced blocks stay upright or remain stacked, whereas in the reference they scatter across the ground. 
The second-round reconstruction matches the reference most closely in both the positions and orientations of the fallen blocks (green boxes). As in the reference, its blocks tip over and spread out across the ground.

\subsubsection{Deformable Object Reconstruction}
\label{sec:exp-deformable}
For deformable objects, the agent reconstructs a soft body from a reference video and writes a program that reproduces its deformation over the whole clip.

Overall, four of the five models achieve better performance on Code2RSI.
The gains appear first in appearance and verifier score.
DeepSeek-V4.1-Flash doubles its DINOv3 from 0.298 to 0.597 and triples its SS score from 1.1 to 3.5.
Similarly, GPT-6-Astra rises from 0.466 to 0.696 in DINOv3 and reaches the highest SS of 4.7.
For these two models all three metrics rise again in the second round, showing that the retained updates accumulate across rounds.
GLM-5.3-Flash gains most in the first round, where DINOv3 rises from 0.239 to 0.366 and SS from 1 to 1.9.
The gains in motion are smaller.
GPT-6-Astra reaches the highest Dynamic IoU of 0.151, and GPT-6 Luna raises it from 0.012 to 0.1 in the second round. Qwen3.8-27B is the only model that falls in both DINOv3 and Dynamic IoU under RSI.
These results show that Code2RSI improves the appearance and the verifier score of soft body reconstruction across backbones and reproducing the deformation remains the harder part of this task.

\subsubsection{Robotic Interaction}
\label{sec:exp-robotic-interaction}
For robotic interaction, the agent generates control programs that act within a given simulated scene to complete manipulation tasks.

Similar to the findings in dynamic scene reconstruction, Code2RSI improves performance across backbone models.
As Table~\ref{tab:real2sim-three-domain-results} shows, for every model at least one RSI round exceeds the fixed harness in progress score. 
GPT-6-Astra reaches the highest progress score of 0.857 after the first round. The improvement differs across models. For example, GPT-6 Luna improves in the first round but falls back in the second.
GLM-5.3-Flash starts from 0.482 under the fixed harness and reaches 0.668 after the second round. 
For DeepSeek-V4.1-Flash, the progress score rises in both rounds, from 0.619 to 0.7 and then to 0.782.

Figure~\ref{fig:robot-pick-keyframes} shows a pick-and-place program generated by Code2RSI. The task requires the arm to move the red cube onto the target marker and release it. The arm first approaches the cube from above and lowers the gripper to grasp it. It then lifts the cube and carries it above the target (1.9 s), before lowering it onto the marker (2.52 s). In the final frame (3.15 s), the gripper has released the cube and moved away, leaving the cube on the target. The program separates transport from placement: the cube is raised during transfer and lowered before release. Appendix~\ref{app:robot} provides further examples of rearrangement, push and pull tasks.

\begin{figure}[!t]
  \centering
  \includegraphics[width=0.99\linewidth]{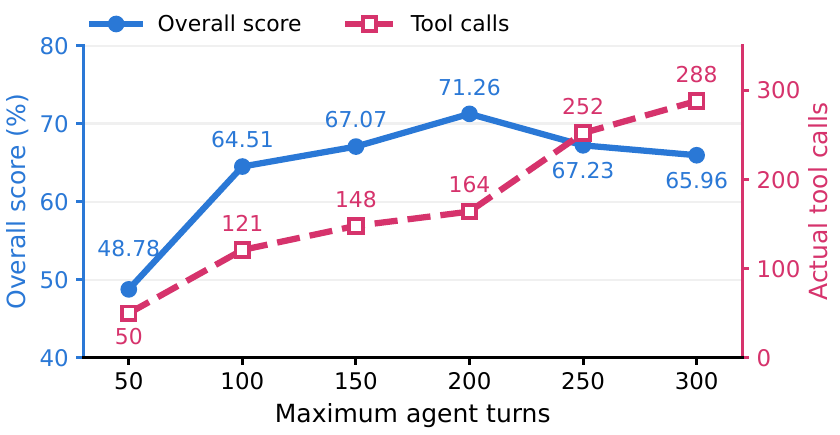}
  \caption{Inner-loop budget ablation with Harness.}
  \label{fig_ablation_budget}
  \centering
  \setlength{\tabcolsep}{2pt}
  \renewcommand{\arraystretch}{1}
  \captionof{table}{Effect of RSI Rounds of Code2RSI.}
  \label{tab:ablation-b02-round-metrics}
  \begin{tabular*}{\columnwidth}{@{\extracolsep{\fill}}lccccc@{}}
    \hline
    Metric & $r=1$ & $r=2$ & $r=3$ & $r=4$ & $r=5$ \\
    \hline
    DINOv3 & 0.842 & 0.810 & 0.867 & \textbf{0.899} & 0.817 \\
    IoU & 0.385 & 0.356 & 0.358 & \textbf{0.508} & 0.323 \\
    Scene 3D & 0.993 & 0.985 & 0.992 & 0.981 & \textbf{0.993} \\
    \hline
  \end{tabular*}
\end{figure}

\begin{table}[!t]
  \centering
  \small
  \setlength{\tabcolsep}{3pt}
  \renewcommand{\arraystretch}{1.12}
  \caption{Component ablation for unseen-task transfer. BH: Base Harness; TG: Transferred Guidance; TS: Transferred Skills; PC: executable Physical Checks in the transferred skills.}
  \label{tab_ablation_components}
  \begin{tabular*}{\columnwidth}{@{\extracolsep{\fill}}ccccccc@{}}
    \hline
    BH & TG & TS & PC & Overall (\%) & Turns & Tools \\
    \hline
    $\checkmark$ & $\times$ & $\times$ & $\times$ & 64.51 & 95 & 121 \\
    $\checkmark$ & $\checkmark$ & $\times$ & $\times$ & 68.82 & 100 & 105 \\
    $\checkmark$ & $\checkmark$ & $\checkmark$ & $\times$ & 67.78 & 99 & 102 \\
    $\checkmark$ & $\checkmark$ & $\checkmark$ & $\checkmark$ & 70.01 & 93 & 95 \\
    \hline
  \end{tabular*}
\end{table}

\subsection{Ablation Study}
\label{sec_ablation}

We select the task of 'pouring liquid from a bottle into a glass' and perform ablation studies of the harness components and RSI loop to evaluate the contribution of different components in Code2RSI.
More detailed settings can be found in Appendix~\ref{app:ablation-details}.

\textbf{Construction Budget of Inner Harness Loop} The default settings in the main experiments allow at most 100 agent turns within the harness. Each turn consists of one agent response, which may request multiple tool calls. 
Figure~\ref{fig_ablation_budget} compares the performance under different turn limits \{50, 150, 200, 250, 300\} within the harness. 
Compared with 64.51\% at the default limit of 100 turns, the reconstruction score drops to 48.78\% at 50 turns and rises to 67.07\% at 150 turns, with the highest score of 71.26\% at 200 turns.
This suggests that increasing the turn budget can lead to more tool calls and better reconstruction within this range. However, additional tool calls lead to lower scores when over 200 turns.
Although tool calls increase from 164 at 200 turns to 252 and 288 when the turn limit increases to 250 and 300 respectively, the overall performance score drops to 67.23\% and 65.96\%. These results show that more tool calls do not necessarily improve reconstruction and suggest that continued revision may move the generated world away from the reference.

\begin{figure}[t]
    \centering
    \includegraphics[width=1\linewidth]{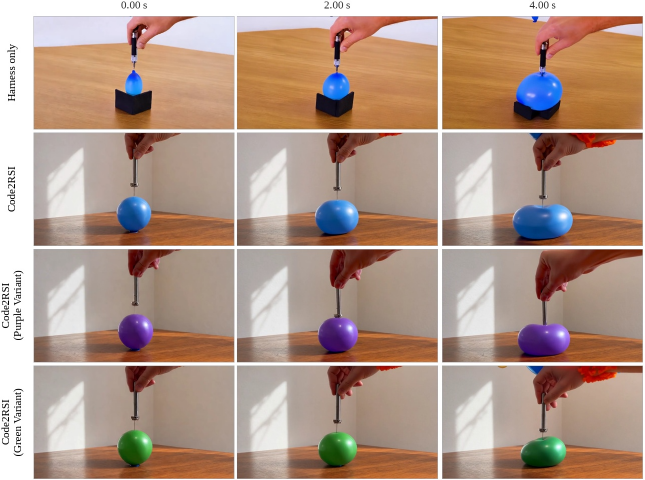}
    \caption{Visual Comparison and Appearance Control on the Balloon Puncture Task. The first two rows compare the fixed harness with Code2RSI. The last two rows only modified the balloon colour in the reference image.}
    \label{fig:phyg-balloon-control}
\end{figure}

\begin{figure*}[t!]
  \centering
  \includegraphics[width=\textwidth]{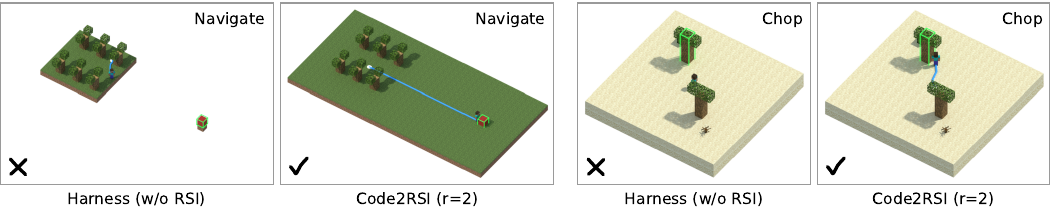}
  \caption{Comparison of Worlds Constructed by the Fixed Harness Only and Code2RSI on Minecraft Navigation and Chop Tasks. Blue lines show player trajectories. Green boxes mark task targets.}
  \label{fig:mc-scenes-v2}
\end{figure*}

\textbf{Rounds and Generalization of Outer RSI Loop} 
To study how the number of RSI rounds affects world reconstruction quality, we compare five rounds of Code2RSI while fixing the inner-loop budget at 100 turns. We evaluate successive updates on the same target task.
Table~\ref{tab:ablation-b02-round-metrics} shows that DINOv3 and IoU peak at $r=4$ while Scene 3D remains high across all rounds. Additional RSI rounds do not always improve reconstruction quality, it shows the trade-off between further refinement and the risk of performance degradation.  
We also evaluate if retained RSI updates transfer beyond the tasks that produced them. As Table~\ref{tab_ablation_components} shows, guidance learned on four source tasks raises the unseen target's score from 64.51\% to 68.82\%. The full combination of guidance, skills and physical checks reaches 70.01\%, with 95 tool calls compared with 121 for the base harness.
Appendix~\ref{app:ablation-details} compares the cost of Harness and Code2RSI and the results show that Code2RSI achieves better performance with a smaller budget.

\textbf{Effect of Physical Checks}
To assess the contribution of physical checks, we disable the gravity, support penetration and rigid shape diagnostics in the skills while keeping the guidance and other skill content unchanged.
We find that physical checks are important to the skill transfer gains. As shown in Table~\ref{tab_ablation_components}, adding skills without these checks lowers the score from 68.82\% with guidance alone to 67.78\%.
Enabling the checks raises the score by 2.23\% while reducing agent turns from 99 to 93 and tool calls from 102 to 95.
The full configuration achieves the highest score with the fewest turns and tool calls among all variants.

\subsection{Application Beyond Real2Sim}
\label{sec:exp-beyond-real2sim}

We extend the application of Code2RSI to text-conditioned scene generation and task-driven game construction to show its potential across different domains. The implementations details and more demos can be found in Appendix~\ref{app:beyond-real2sim}.

\textbf{Text-Conditioned Scene Generation}
\label{sec:exp-text-scene}
For text-conditioned scene generation, the model writes a scene program from the textual description. The program defines the geometry and animation used to guide the final video generation. We evaluate the generated videos with VBench~\cite{huang2024vbench}.
As Table~\ref{tab:beyond-results} shows, the first RSI round raises imaging quality from 0.647 to 0.668. Subject consistency and motion smoothness change little across rounds. 
Figure~\ref{fig:phyg-balloon-control} compares the generated videos on the balloon puncture task. Under the fixed harness, the balloon keeps growing and its stand changes shape. With Code2RSI, the needle presses into the balloon and the surrounding scene remains more stable. The last two rows show how the reference image controls appearance. Only the balloon colour in the reference image changes, while the world program remains fixed. These examples demonstrate promising qualitative results when applying Code2RSI to program-guided visual generation, despite the inconclusive aggregate metrics. Appendix~\ref{app:rendering} provides further examples and detailed results.

\begin{table}[t]
  \caption{Results Beyond Real2Sim Using DeepSeek-V4.1-Flash. SC: subject consistency; IQ: imaging quality; MS: motion smoothness; TC: task completion; ST: static score.}
  \label{tab:beyond-results}
  \centering
  \small
  \setlength{\tabcolsep}{2pt}
  \begin{tabular*}{\columnwidth}{@{\extracolsep{\fill}}lccccc@{}}
    \toprule
    & \multicolumn{3}{c}{Text-to-Scene} & \multicolumn{2}{c}{Gaming} \\
    \cmidrule(lr){2-4} \cmidrule(l){5-6}
    Method & SC $\uparrow$ & IQ $\uparrow$ & MS $\uparrow$ & TC $\uparrow$ & ST $\uparrow$ \\
    \midrule
    Harness (w/o RSI) & 0.931 & 0.647 & 0.992 & 0.417 & 82.8 \\
    Code2RSI ($r{=}1$) & 0.925 & 0.668 & 0.990 & 0.667 & 100.0 \\
    Code2RSI ($r{=}2$) & 0.935 & 0.644 & 0.994 & 0.750 & 100.0 \\
    \bottomrule
  \end{tabular*}
\end{table}

\textbf{Task-Driven Game Construction}
\label{sec:exp-game-construction}
For task-driven game construction, the agent writes programs that build a Minecraft scene and supply the resources required by the task. A player agent then attempts the task in the constructed world. Task completion measures the player's success in achieving the goal. The static score measures program execution and the presence of required content.
Table~\ref{tab:beyond-results} shows that task completion rises from 0.417 under the fixed harness to 0.667 after the first RSI round. The second round further raises it to 0.75. The static score reaches 100\% after the first round and remains unchanged in the second.
Figure~\ref{fig:mc-scenes-v2} shows how the constructed layout affects a navigation task in Minecraft. The fixed harness builds ground that ends before the red target. Code2RSI extends the ground and the recorded player trajectory reaches the target. The retained guidance tells the constructor to place the spawn above ground and provide reachable paths to workstations. This example illustrates that Code2RSI can translate task-level feedback into reusable guidance for improving the functionality of generated environments. 

These results highlight the broader applicability of Code2RSI across different domains, showing promising qualitative results in visual generation and substantial improvements in downstream interactive environments.

\section{Conclusion}

We presented Code2RSI, an agentic framework that organizes world construction through task-specific agent harnesses with a focus on Real2Sim.
Each harness connects world programs with execution tools, reliability checks and feedback. The inner loop revises the current world program, and the outer loop retains updates to reusable construction skills and harness routines.
The results show that retained rounds can improve dynamic scene reconstruction and task progress in robotic interaction. Extensions to text-conditioned scene generation and task-driven game construction also show promising results beyond Real2Sim.
These results highlight the potential of execution-driven recursive self-improvement and motivate further applications in reliable world construction.

\clearpage

\section*{LLM Usage Statement}
We used OpenAI Codex to assist with developing and refining experimental ideas, implementing and revising research code, and drafting and editing manuscript text. The authors defined the research objectives, led the methodological and experimental design, and analyzed and interpreted the results. Codex-generated suggestions, code, and text were evaluated and revised as appropriate by the authors, who determined whether and how to incorporate them into the work. The authors take full responsibility for the final manuscript and the reported findings. The LLM backbones used in Code2RSI and the image and video generation models are documented in the main text and supplementary material.
{
    \small
    \bibliographystyle{ieeenat_fullname}
    \bibliography{main}
}

\clearpage
\setcounter{page}{1}
\maketitlesupplementary
\appendix

\begin{table*}[t!]
  \centering
  \footnotesize
  \setlength{\tabcolsep}{4pt}
  \renewcommand{\arraystretch}{1.15}
  \caption{Implementation of Code2RSI Across Experimental Settings.}
  \label{tab:domain-setup}
  \begin{tabular*}{\textwidth}{@{\extracolsep{\fill}}ccccc@{}}
    \hline
    Setting & Task source ($T$) & World program ($P$) & Execution backend ($E$) & Verification ($V$) \\
    \hline
    Dynamic scene & 4DCodeBench & Scene script & Blender & Reconstruction metrics \\
    Deformable object & PhysTwin & Deformation script & Simulator & Verifier and visual metrics \\
    Robotic interaction & PyBullet tasks & Control function & PyBullet & Completion and visual metrics \\
    Text-to-scene & PhyGenBench & Blender script & Blender + image/video models & Geometry and video quality \\
    Minecraft & MineDojo tasks & Command files & Minecraft + player agent & Scene and gameplay checks \\
    \hline
  \end{tabular*}
\end{table*}

\section{Experiments on the Real2Sim Domain}
\label{app:real2sim}

\subsection{Methodology}
\label{app:real2sim-metrics}

\textbf{Metrics}

For dynamic scene reconstruction, we follow the evaluation definitions of 4DCodeBench~\cite{shen2026fourdcodebench}. We compare the reference and rendered videos at $T$ sampled frames indexed by $t$. Reference quantities are unmarked, while reconstructed quantities carry a hat. All five metrics lie in $[0,1]$, with higher scores indicating better agreement.

\emph{DINOv3} measures appearance by the mean cosine similarity between the DINOv3~\cite{simeoni2025dinov3} class-token embeddings of corresponding frames, mapped to $[0,1]$. Let $e_t$ and $\hat e_t$ denote the embeddings at frame $t$:
\begin{equation}
  \mathrm{DINOv3}=\frac{1}{2}\Big(1+\frac{1}{T}\sum_{t=1}^{T}\frac{e_t^{\top}\hat e_t}{\|e_t\|\,\|\hat e_t\|}\Big).
\end{equation}

\emph{IoU} uses task-specific mask definitions and aggregation. For deformable object reconstruction, it measures the pooled overlap of RAFT-derived visible-motion masks. For dynamic scene reconstruction, $\mathrm{DynIoU}=\frac{1}{T}\sum_{t=1}^{T}\frac{|(M_t\cap\hat M_t)\setminus B_t|}{|(M_t\cup\hat M_t)\setminus B_t|}$ follows 4DCodeBench~\cite{shen2026fourdcodebench} and measures image-space overlap of moving objects. Let $M_t$ and $\hat M_t$ denote their masks in the reference and reconstruction, and let $B_t$ be a thin band along the boundary of $M_t$. 
The boundary band is excluded from the intersection and union.
For robotic interaction, DINOv3 and IoU compare each execution with a successful scripted reference demonstration in the same scene and camera view. Frames are matched at identical simulation times, with shorter executions padded by their settled final frames, IoU uses simulator segmentation masks of movable robot and drawer links and dynamic objects.

The remaining three metrics compare 3D geometry and motion after similarity alignment with trimmed ICP. Distances are normalized by the RMS radius of the reference surface at frame 0.

\emph{Scene 3D} measures static geometry using the symmetric Chamfer distance $D_\kappa$ between the reference and reconstructed surfaces visible through the reference camera at frame 0. Per-point distances are capped at $\kappa=0.5$; $\mathcal{P}$ and $\hat{\mathcal{P}}$ denote the corresponding point sets:
\begin{equation}
  \mathrm{Scene3D}=1-\frac{D_\kappa(\mathcal{P},\hat{\mathcal{P}})}{\kappa}.
\end{equation}

\emph{Trajectory DTW} measures motion by comparing material-point trajectories. We track sampled points over time and match the reference and reconstructed trajectories by DTW distance using the Hungarian algorithm. Let $(\tau_i,\hat\tau_i)$ denote the $i$-th matched pair among $n$ pairs; with $\kappa=1$:
\begin{equation}
  \mathrm{TrajDTW}=1-\frac{1}{n}\sum_{i=1}^{n}\frac{\min\big(\mathrm{DTW}(\tau_i,\hat\tau_i),\kappa\big)}{\kappa}.
\end{equation}

\emph{EMD step} compares the unordered sets of per-frame 3D displacements, $U_t$ and $\hat U_t$, using the sliced Wasserstein distance $\mathrm{SW}$. Let $m(\cdot)$ denote the root-mean-square displacement magnitude:
\begin{equation}
  \mathrm{EMD}=\max\Big(0,\;1-\frac{\sum_{t}\mathrm{SW}(U_t,\hat U_t)}{\sum_{t}\big(m(U_t)+m(\hat U_t)\big)}\Big).
\end{equation}

\emph{Overall} is the score of a run. It is the mean of the five metrics in percent. Reported Overall (\%) values are obtained by multiplying the normalized score by 100.
\begin{equation}
\begin{split}
  \mathrm{Overall}=\,1/5 * \big(&\mathrm{DINOv3}+\mathrm{DynIoU}+\mathrm{Scene3D}\\
  &+\mathrm{TrajDTW}+\mathrm{EMD}\big).
\end{split}
\end{equation}

\emph{SS} for deformable object reconstruction is a $0$-$10$ self-score from the generation backbone acting as the VIGA verifier~\cite{yin2026viga}, prompted with reference/rendered keyframes, the task description, generated code and physical observations to assess geometry, appearance, camera, contacts, deformation, motion and task outcome and return the score.

\textbf{Framework Scope and Integration.} Code2RSI provides an integrated framework for executable world construction, evaluation, and recursive self-improvement across task domains. The overall framework is inspired by PenguinHarness. Our contribution centers on organizing domain-specific world representations, execution backends, observations, and verification procedures into a shared construction workflow. Each domain instantiates this workflow through a task-specific harness that connects the coding agent to the tools and execution evidence needed to construct and assess its world programs. This organization allows the same construction and self-improvement procedure to operate across tasks with different program representations and criteria for success.
Harness (w/o RSI) is an ablation that retains inner-loop program refinement while keeping the initial harness fixed. RSI develops reusable guidance and skills on a training set, and evaluates the retained configuration on held-out test tasks of the same task types. A candidate update is accepted only if it improves the mean training score over the current retained configuration. Otherwise, the latest configuration is preserved with up to two retries allowed. Harness updates are selected under fixed evaluation criteria and inner-loop budgets. So the performance on held-out tasks may not improve monotonically across RSI rounds. Successive harness configurations are evolved on the training set and evaluated on the same held-out evaluation task. Inner-loop program refinement remains enabled during evaluation. Executable physical checks are evolvable diagnostics within harness skills, while the evaluation verifier remains fixed.

\subsection{Dynamic Scene Construction Domain}
\label{app:dynamic-scene}

\begin{figure*}
    \centering
    \includegraphics[width=1\linewidth]{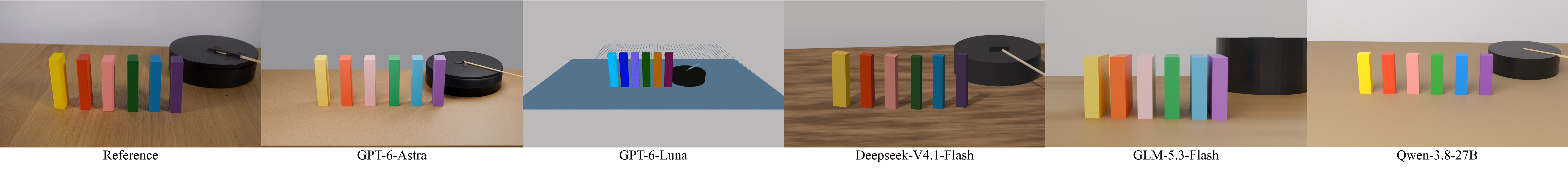}
    \caption{Visual Comparison of a Dynamic Scene Across Models.}
    \label{fig:backbone-reconstructions}
\end{figure*}

Table~\ref{tab:rsi-iteration-rounds-luna} reports the five reconstruction metrics and their aggregate score for each backbone across the initial harness and two RSI rounds. \textcolor{black}{GPT-6-Astra} improves from 66.67\% to 71.8\%, while \textcolor{black}{GPT-6 Luna} improves from 25.1\% to 51.91\% after an initial decline. \textcolor{black}{DeepSeek-V4.1-Flash} peaks at 65.04\% in the first round and falls to 59.33\% in the second, illustrating that additional rounds do not guarantee improvement.

\begin{table*}[t!]
  \caption{Reconstruction Performance Across RSI Rounds on Dynamic Scene Reconstruction.}
  \label{tab:rsi-iteration-rounds-luna}
  \centering
  \small
  \setlength{\tabcolsep}{5pt}
  \renewcommand{\arraystretch}{1.18}
  \begin{tabular*}{\textwidth}{@{\extracolsep{\fill}}lrrrrrr@{}}
    \toprule
    Round & DINOv3 $\uparrow$ & Dyn. IoU $\uparrow$ & Scene 3D $\uparrow$ & Traj. DTW $\uparrow$ & EMD step $\uparrow$ & Overall (\%) $\uparrow$ \\
    \midrule
    \multicolumn{7}{@{}l}{\textit{GPT-6-Astra}} \\
    Harness (w/o RSI) & 0.898 & 0.355 & \textbf{0.880} & 0.734 & \textbf{0.468} & 66.67 \\
    Code2RSI ($r{=}1$) & 0.884 & 0.459 & 0.836 & \textbf{0.908} & 0.351 & 68.72 \\
    Code2RSI ($r{=}2$) & \textbf{0.907} & \textbf{0.584} & 0.868 & 0.901 & 0.332 & \textbf{71.8} \\
    \midrule
    \multicolumn{7}{@{}l}{\textit{GPT-6 Luna}} \\
    Harness (w/o RSI) & 0.450 & 0.000 & 0.449 & 0.177 & 0.179 & 25.1 \\
    Code2RSI ($r{=}1$) & 0.436 & 0.000 & 0.457 & 0.263 & 0.067 & 24.47 \\
    Code2RSI ($r{=}2$) & \textbf{0.871} & \textbf{0.071} & \textbf{0.836} & \textbf{0.589} & \textbf{0.229} & \textbf{51.91} \\
    \midrule
    \multicolumn{7}{@{}l}{\textit{DeepSeek-V4.1-Flash}} \\
    Harness (w/o RSI) & 0.428 & 0.208 & 0.427 & 0.408 & 0.174 & 32.89 \\
    Code2RSI ($r{=}1$) & \textbf{0.841} & \textbf{0.360} & \textbf{0.860} & 0.752 & \textbf{0.438} & \textbf{65.04} \\
    Code2RSI ($r{=}2$) & 0.828 & 0.142 & 0.854 & \textbf{0.829} & 0.314 & 59.33 \\
    \midrule
    \multicolumn{7}{@{}l}{\textit{GLM-5.3-Flash}} \\
    Harness (w/o RSI) & 0.429 & 0.000 & 0.129 & 0.000 & 0.000 & 11.17 \\
    Code2RSI ($r{=}1$) & \textbf{0.846} & 0.001 & 0.369 & \textbf{0.467} & 0.113 & \textbf{35.9} \\
    Code2RSI ($r{=}2$) & 0.414 & \textbf{0.136} & \textbf{0.419} & 0.430 & \textbf{0.152} & 31.02 \\
    \midrule
    \multicolumn{7}{@{}l}{\textit{Qwen3.8-27B}} \\
    Harness (w/o RSI) & 0.254 & 0.000 & 0.404 & 0.000 & 0.000 & 13.15 \\
    Code2RSI ($r{=}1$) & 0.739 & \textbf{0.217} & 0.718 & 0.579 & 0.188 & 48.81 \\
    Code2RSI ($r{=}2$) & \textbf{0.868} & 0.033 & \textbf{0.835} & \textbf{0.637} & \textbf{0.264} & \textbf{52.75} \\
    \bottomrule
  \end{tabular*}
\end{table*}

\subsection{Robotic Interaction Domain}
\label{app:robot}

The robot world program is a control script that moves a robot arm inside a built scene. The harness executes the script in a physics simulator, records a video and reads the final state of the scene. Verification compares this final state with the completion condition of the task.

\textbf{Settings.} The robot tasks are built in PyBullet~\cite{coumans2016pybullet}. 
Each task includes a textual objective and a scene of coloured blocks, goal markers and a table in front of a Franka Panda arm. We build 7 types of tasks including pick-and-place, rearrangement, push, pull, packing, assembly and open drawer. 
Table~\ref{tab:robot-tasks} lists the objective and the completion condition of each type. The task completion rate (TC) of a set of tasks is the ratio of completed tasks.
For example, the push task asks the robot to pick up a dark paddle, push the red block onto the green marker by contact without lifting or grasping the block, and return the paddle to its start position.
The world program $P$ is a Python function that receives the scene context and controls the robot. The context exposes the observed entities with positions, shapes and colours, the simulator client and the robot controller with move, gripper and grasp commands.
The backend $E$ is a headless PyBullet process. It builds the scene, runs the function, records the video and measures the result. DeepSeek-V4.1-Flash is set as the default LLM for both policy generation and optimization.

\textbf{Metrics.} We report task completion (TC) and progress score (PS) across tasks. TC is the task completion rate defined above. TC measures goal-state completion, while physical validity is evaluated separately through execution diagnostics. PS is the mean of a per-task score in $[0, 1]$. Specifically, a completed task scores 1, and a failed task receives a score reflecting how much of the task it has accomplished. The scoring rule depends on the task’s completion condition, using measures such as the distance to a goal or the fraction of objects correctly placed. Scores are reduced if the robot has not released its grasp, and failed tasks can score no higher than 0.99.
The per-task columns of Table~\ref{tab:robot-detail} list PS for each task type, and a value of a smaller value shows how far the program got.

\textbf{Results.} 
As the results in Table~\ref{tab:robot-detail} show, the harness alone reaches a TC of 57.1\% and a PS of 61.9\%. It passes tasks of pick-and-place, rearrangement, push and pull, and fails packing, assembly and open drawer.
Round 1 Code2RSI keeps TC at 57.1\% and raises PS to 70\%.
The updated harness covers classifying entities by role instead of by exclusion, re-observing the scene after every motion primitive and reading the joint axis of the drawer before pulling. 
Round 2 raises TC to 71.4\% and PS to 78.2\%. The updated harness adds placement inside the container interior computed from its bounding box, a pull procedure that keeps the block on the table and an assembly order that settles the pillars before the beam.

\begin{table*}[t]
  \caption{Task Types and Completion Conditions in the Robot Domain.}
  \label{tab:robot-tasks}
  \centering
  \small
  \setlength{\tabcolsep}{6pt}
  \renewcommand{\arraystretch}{1.15}
  \begin{tabular}{@{}c>{\centering\arraybackslash}m{9cm}>{\centering\arraybackslash}m{6cm}@{}}
    \hline
    Task Type & Objective & Completion condition \\
    \hline
    Pick-and-place & Pick up the cube, place it on the target marker and release it. & Cube within 0.05 m of the marker. \\
    Rearrangement & Move each cube to the marker of the same colour and release them. & Every cube within 0.05 m of its marker. \\
    Assembly & Place the pillars upright on their markers, then lay the beam across their tops so it rests on both. & Both pillars on their markers and the beam resting on both. \\
    Push & Pick up the paddle and use it to push the block onto the marker, then return the paddle to its start. & Block within 0.05 m of the marker and the paddle returned to its start. \\
    Pull & Grasp the block and pull it along the table onto the marker. & Block within 0.05 m of the marker. \\
    Packing & Pick up the blocks and pack all of them inside the open box. & Bounding box of every block inside the box. \\
    Open drawer & Grasp the handle and pull the drawer open along its sliding axis, leaving part of the tray inside the cabinet. & Drawer opening inside its allowed range. \\
    \hline
  \end{tabular}
\end{table*}

\begin{table*}[t]
  \caption{Task-Level Results in the Scenes of the Robot Interaction. Each task cell reports the progress score under the given task.}
  \label{tab:robot-detail}
  \centering
  \small
  \setlength{\tabcolsep}{4pt}
  \renewcommand{\arraystretch}{1.15}
  \begin{tabular*}{\textwidth}{@{\extracolsep{\fill}}cccccccccc@{}}
    \toprule
    Method & TC & PS & Pick & Rearrange & Push & Pull & Packing & Assembly & Drawer \\
    \midrule
    \multicolumn{10}{@{}l}{\textit{GPT-6-Astra}} \\
    Harness (w/o RSI) & 0.714 & 0.814 & 1.000 & 1.000 & 0.513 & 1.000 & 1.000 & 1.000 & 0.182 \\
    Code2RSI ($r{=}1$) & 0.857 & 0.857 & 1.000 & 1.000 & 1.000 & 1.000 & 0.000 & 1.000 & 1.000 \\
    Code2RSI ($r{=}2$) & 0.714 & 0.856 & 1.000 & 1.000 & 1.000 & 1.000 & 0.000 & 1.000 & 0.990 \\
    \midrule
    \multicolumn{10}{@{}l}{\textit{GPT-6 Luna}} \\
    Harness (w/o RSI) & 0.429 & 0.756 & 1.000 & 0.803 & 1.000 & 1.000 & 0.000 & 0.500 & 0.990 \\
    Code2RSI ($r{=}1$) & 0.429 & 0.808 & 1.000 & 1.000 & 0.500 & 1.000 & 0.667 & 0.500 & 0.990 \\
    Code2RSI ($r{=}2$) & 0.571 & 0.619 & 1.000 & 1.000 & 1.000 & 1.000 & 0.333 & 0.000 & 0.000 \\
    \midrule
    \multicolumn{10}{@{}l}{\textit{DeepSeek-V4.1-Flash}} \\
    Harness (w/o RSI) & 0.571 & 0.619 & 1.000 & 1.000 & 1.000 & 1.000 & 0.333 & 0.000 & 0.000 \\
    Code2RSI ($r{=}1$) & 0.571 & 0.700 & 1.000 & 0.240 & 0.660 & 1.000 & 1.000 & 1.000 & 0.000 \\
    Code2RSI ($r{=}2$) & 0.714 & 0.782 & 0.475 & 1.000 & 1.000 & 1.000 & 1.000 & 0.000 & 1.000 \\
    \midrule
    \multicolumn{10}{@{}l}{\textit{GLM-5.3-Flash}} \\
    Harness (w/o RSI) & 0.286 & 0.482 & 0.663 & 1.000 & 0.660 & 1.000 & 0.000 & 0.000 & 0.053 \\
    Code2RSI ($r{=}1$) & 0.286 & 0.471 & 0.219 & 1.000 & 0.660 & 1.000 & 0.000 & 0.000 & 0.420 \\
    Code2RSI ($r{=}2$) & 0.571 & 0.668 & 0.111 & 1.000 & 0.558 & 1.000 & 1.000 & 1.000 & 0.004 \\
    \midrule
    \multicolumn{10}{@{}l}{\textit{Qwen3.8-27B}} \\
    Harness (w/o RSI) & 0.286 & 0.493 & 0.000 & 1.000 & 0.941 & 1.000 & 0.000 & 0.000 & 0.510 \\
    Code2RSI ($r{=}1$) & 0.571 & 0.571 & 1.000 & 1.000 & 1.000 & 1.000 & 0.000 & 0.000 & 0.000 \\
    Code2RSI ($r{=}2$) & 0.571 & 0.713 & 1.000 & 1.000 & 1.000 & 1.000 & 0.000 & 0.000 & 0.990 \\
    \bottomrule
  \end{tabular*}
\end{table*}

\begin{figure}[t]
  \centering
  \includegraphics[width=0.5\textwidth]{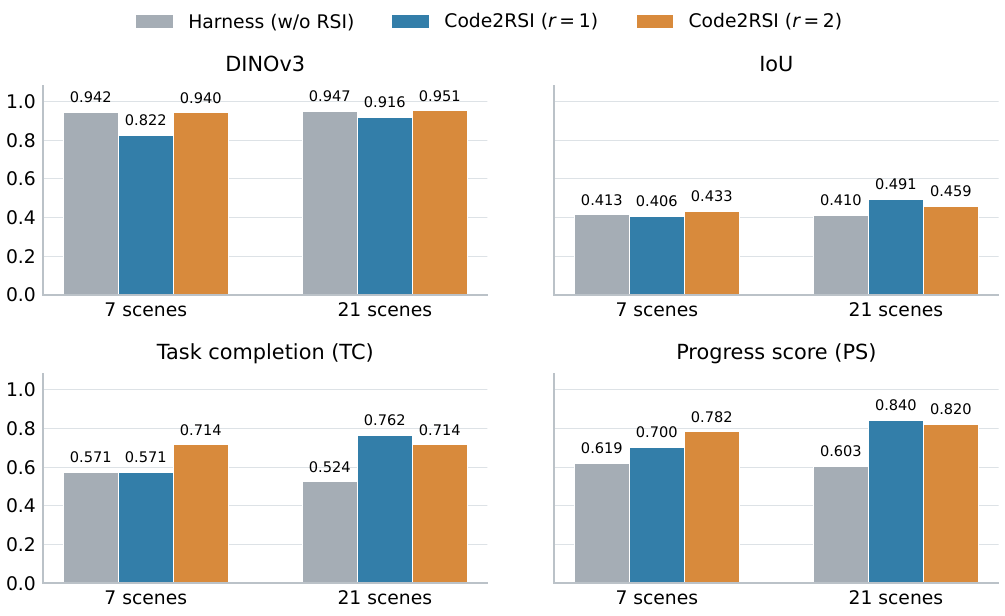}
  \caption{Performance on the 7 original robot scenes and the extended set of 21 scenes. Higher values indicate better performance for all metrics.}
  \label{fig:robot-deepseek-test-size}
\end{figure}

\textbf{Generalization to Additional Robot Scenes}
\label{app:robot-test-size}
To assess generalization to additional scenes, we expand the robot test set from 7 to 21 scenes and evaluate their performance on DeepSeek-V4.1-Flash.
As Figure~\ref{fig:robot-deepseek-test-size} shows, both evaluated RSI stages achieve better performance on the expanded set.
This is consistent with our earlier observation that further RSI does not guarantee better performance and an earlier retained state can generalize better.

\begin{figure*}[!htbp]
  \centering
  \includegraphics[width=\textwidth]{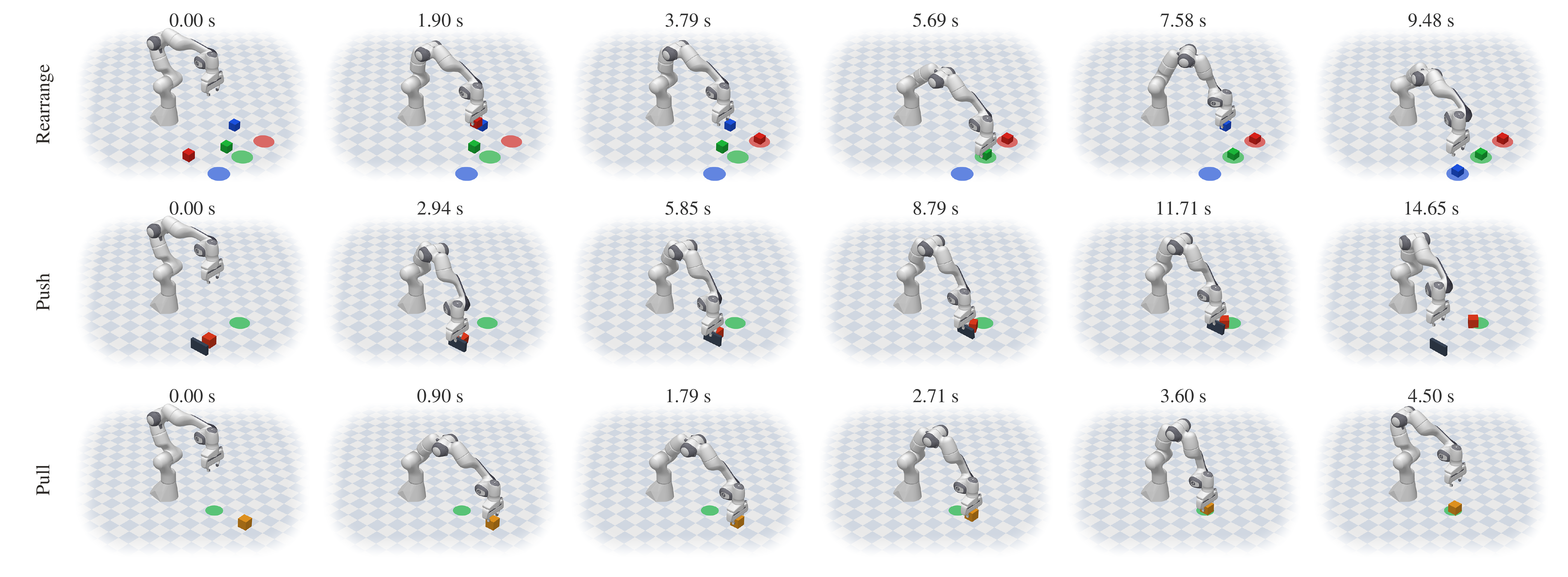}
  \caption{Keyframes of Code2RSI's Generated Policies ($r{=}2$) on Rearrangement, Push and Pull Tasks. Top: move each cube to the marker of the same colour and release it. Middle: pick up the paddle, push the block onto the marker, and return the paddle to its start. Bottom: grasp the block and pull it along the table onto the marker.}
  \label{fig:supp-robot-rsi-keyframes}
\end{figure*}

\textbf{Additional Keyframes of Robot Scenes.} Figure~\ref{fig:supp-robot-rsi-keyframes} shows the keyframes of rearrangement, push and pull tasks in robotic interaction. Each sequence shows six evenly spaced keyframes from a saved policy execution, with simulation time above each frame. The policies are generated by Code2RSI with DeepSeek-V4.1-Flash at $r{=}2$.

\subsection{Deformable Object Reconstruction}

\begin{figure*}[t]
  \centering
  \includegraphics[width=\textwidth]{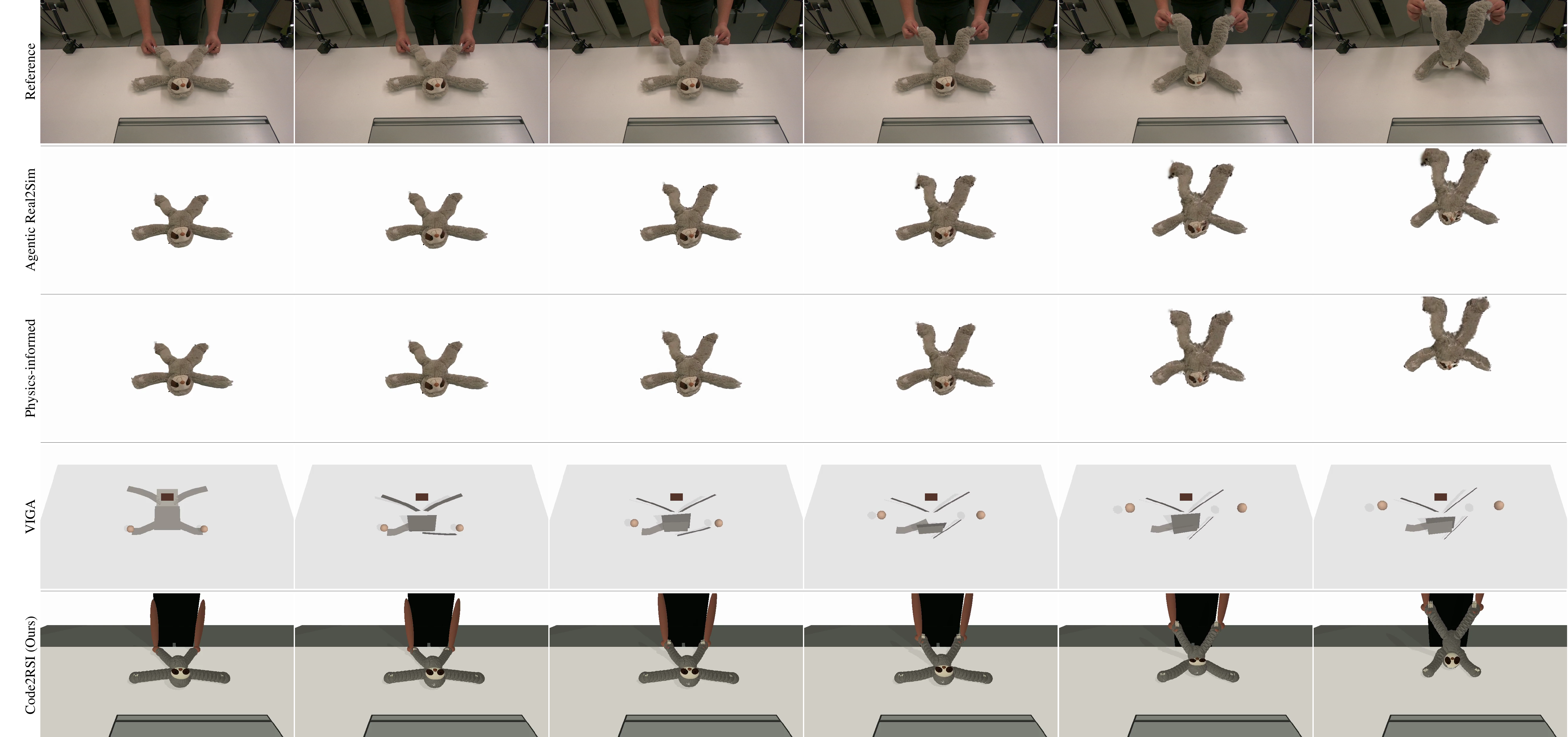}
  \caption{Visual Comparison on the Task of Lifting a Sloth Toy with Both Hands. From top to bottom: reference video, Agentic Real2Sim, Physics-informed, VIGA and Code2RSI. Each row contains six uniformly sampled frames.}
  \label{fig:supp-deformable-double-lift-sloth}
\end{figure*}

\begin{figure*}[t]
  \centering
  \includegraphics[width=\textwidth]{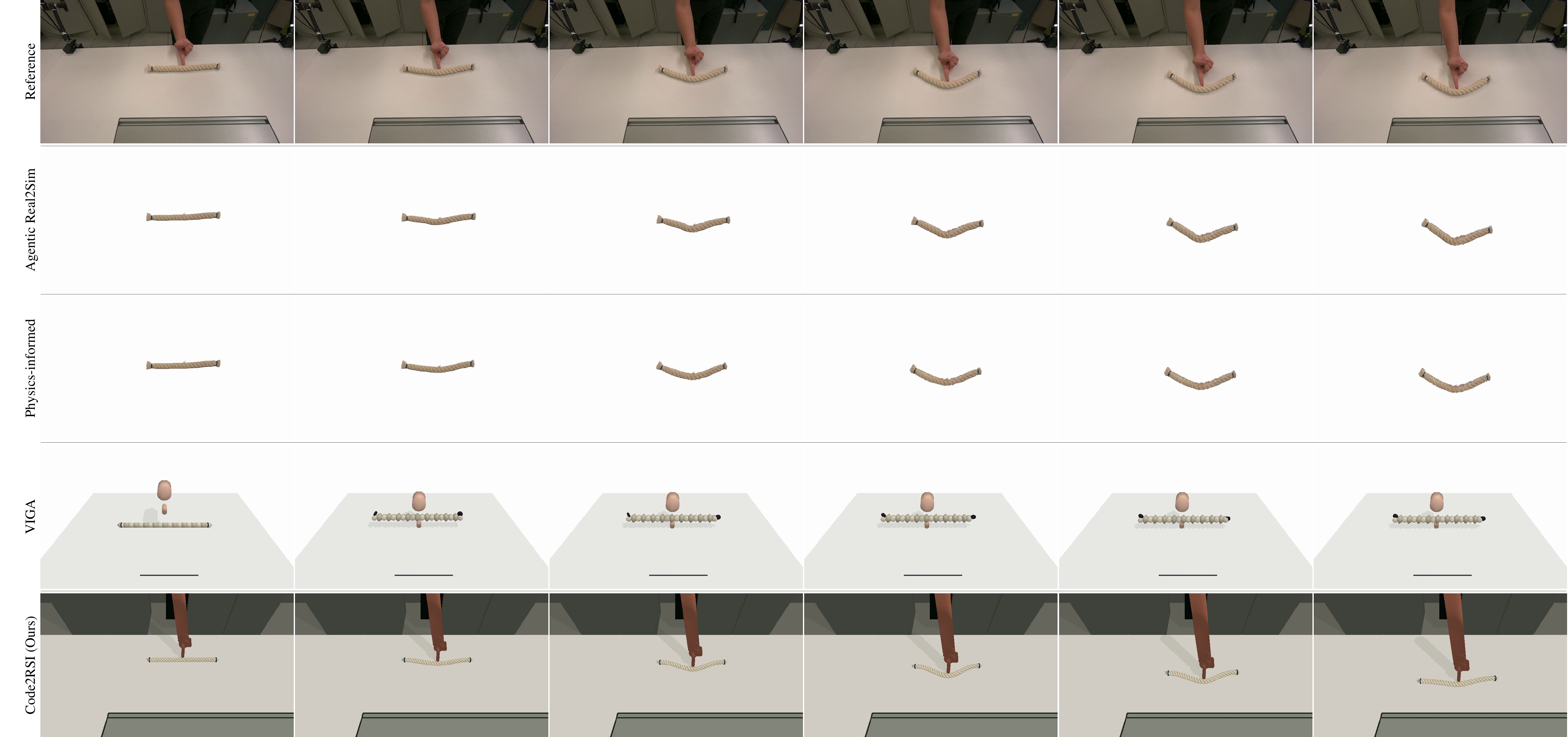}
  \caption{Visual Comparison on the Task of Pushing a Rope with One Hand. From top to bottom: reference video, Agentic Real2Sim, Physics-informed, VIGA and Code2RSI. Each row contains six uniformly sampled frames.}
  \label{fig:supp-deformable-single-push-rope}
\end{figure*}

Figures~\ref{fig:supp-deformable-single-push-rope} and~\ref{fig:supp-deformable-double-lift-sloth} compare the visual reconstruction performance of Code2RSI and different baselines on two deformable object reconstruction tasks: `Pushing a rope with one hand' and `Lifting a sloth toy with both hands', respectively. The baselines include Agentic Real2Sim~\cite{chen2026agenticreal2sim}, the Physics-informed baseline~\cite{jiang2025phystwin}, and VIGA~\cite{yin2026viga}. Each figure shows six uniformly sampled frames from the real reference video and each method's reconstruction.
Within each setting, execution feedback supports revision of the current program, while construction records inform updates to reusable modeling guidance and harness skills. To conclude, Code2RSI integrates world generation and evaluation with improvement of the construction process itself. 

\subsection{Case Study}
\label{app:ablation-details}

\textbf{Construction Budget}
We further compare the construction cost of Harness Only and Code2RSI in tool calls and token usage on the dynamic-scene tasks. Table~\ref{tab:construction-call-accounting} sums tasks per backbone and round. 
Compared with Harness Only, Code2RSI reduces total tool calls by 41.4\% and 36.4\% after one and two rounds, and total tokens by 57.5\% and 41.3\%. After two rounds, all five backbones improve trajectory DTW. Two recorded DeepSeek fixed-harness restarts consume another 257 calls but time out without scored outputs. Longer ordinary revision can attain comparable quality, using 164 calls for 71.26\%, whereas transferred RSI reaches 70.01\% with 95 calls. These results show that Code2RSI can achieve better reconstruction performance with fewer construction calls and tokens.

\begin{table}[t]
\centering\small
\caption{Construction Cost on Dynamic Scenes.}
\label{tab:construction-call-accounting}
\setlength{\tabcolsep}{4pt}
\begin{tabular*}{\columnwidth}{@{\extracolsep{\fill}}lrrr@{}}
\toprule
Method & Tool calls & Input (M) & Output (k) \\
\midrule
\multicolumn{4}{c}{\textit{GPT-6 Astra}} \\
Harness Only & 46 & 0.4 & 15 \\
Code2RSI ($r{=}1$) & 49 & 0.5 & 23 \\
Code2RSI ($r{=}2$) & 37 & 0.4 & 18 \\
\midrule
\multicolumn{4}{c}{\textit{GPT-6 Luna}} \\
Harness Only & 20 & 0.1 & 6 \\
Code2RSI ($r{=}1$) & 27 & 0.2 & 8 \\
Code2RSI ($r{=}2$) & 17 & 0.1 & 8 \\
\midrule
\multicolumn{4}{c}{\textit{DeepSeek-V4.1-Flash}} \\
Harness Only & 587 & 48.1 & 547 \\
Code2RSI ($r{=}1$) & 221 & 15.6 & 198 \\
Code2RSI ($r{=}2$) & 214 & 15.0 & 204 \\
\midrule
\multicolumn{4}{c}{\textit{GLM-5.3-Flash}} \\
Harness Only & 171 & 5.4 & 115 \\
Code2RSI ($r{=}1$) & 189 & 6.6 & 105 \\
Code2RSI ($r{=}2$) & 175 & 7.6 & 150 \\
\midrule
\multicolumn{4}{c}{\textit{Qwen3.8-27B}} \\
Harness Only & 256 & 25.8 & 362 \\
Code2RSI ($r{=}1$) & 147 & 10.9 & 210 \\
Code2RSI ($r{=}2$) & 244 & 23.6 & 287 \\
\midrule
\multicolumn{4}{c}{\textit{Total}} \\
Harness Only & 1080 & 79.8 & 1046 \\
Code2RSI ($r{=}1$) & 633 & 33.9 & 544 \\
Code2RSI ($r{=}2$) & 687 & 46.8 & 667 \\
\bottomrule
\end{tabular*}
\end{table}

\section{Experiments Beyond Real2Sim}
\label{app:beyond-real2sim}

This section provides additional settings and results for the text-conditioned game scene generation, rendering applications and a garment-manipulation study that evaluates task completion and physical consistency.

\subsection{Experimental Setups}
\label{sec:setup}

In the gaming domain the agent writes Minecraft command files that build the scene, supply resources and specify a task from MineDojo, and a player agent then acts toward the task goal.
In the rendering domain the agent writes a Blender script that builds gray geometry, and the rendered gray video conditions a video generation model.
In the robot domain the agent writes a control function for a Franka Panda arm in PyBullet across seven manipulation types.
In the real2sim domain the agent writes a build script that reconstructs a dynamic scene from a reference video. Table~\ref{tab:domain-setup} summarizes the experimental implementations. %

Task completion (TC) is the ratio of tasks whose completion condition is met. The gaming domain adds a static score (SS) of program execution and requested content. The robot domain adds a progress score (PS) with task-specific partial credit based on progress toward each task's completion condition. The rendering task reports VBench subject consistency (SC), imaging quality (IQ) and motion smoothness (MS) of the final videos, and the real2sim domain reports the five official metrics of 4DCodeBench and their mean.

\subsubsection{Gaming Domain}
Table~\ref{tab:gaming-eval} shows that the harness alone completes 0.417 of the tasks. Three of its programs miss a required file and cannot be loaded. One round of Code2RSI raises completions to 0.667 and every program passes the static check. The retained guidance covers the required files, a spawn position above the ground, and a reachable path from the spawn to every workstation. Figure~\ref{fig:mc-scenes-v2} compares the constructed worlds. In jungle navigation, the harness-only program successfully builds the scene but ends the ground before the red target. Code2RSI extends the ground, and the trajectory reaches the target. In desert chop, both programs build the same world, but the harness-only one didn't finish the task, while the trajectory from Code2RSI walks to a tree and chops it.

\subsubsection{Embodied Manipulation Domain}
\label{sec:supp-physical}

We evaluate garment manipulation through generated motion programs, measuring task completion and physical consistency.
Appendix~\ref{app:embodied-results} describes the implementation, evaluation metrics, and results.

\subsubsection{Rendering Domain}
In the rendering application, Blender scene programs define geometry and animation, and the rendered gray videos condition final video generation. Appendix~\ref{app:rendering} describes the construction and rendering pipeline.

\begin{figure*}[t!]
  \centering
  \includegraphics[width=\textwidth]{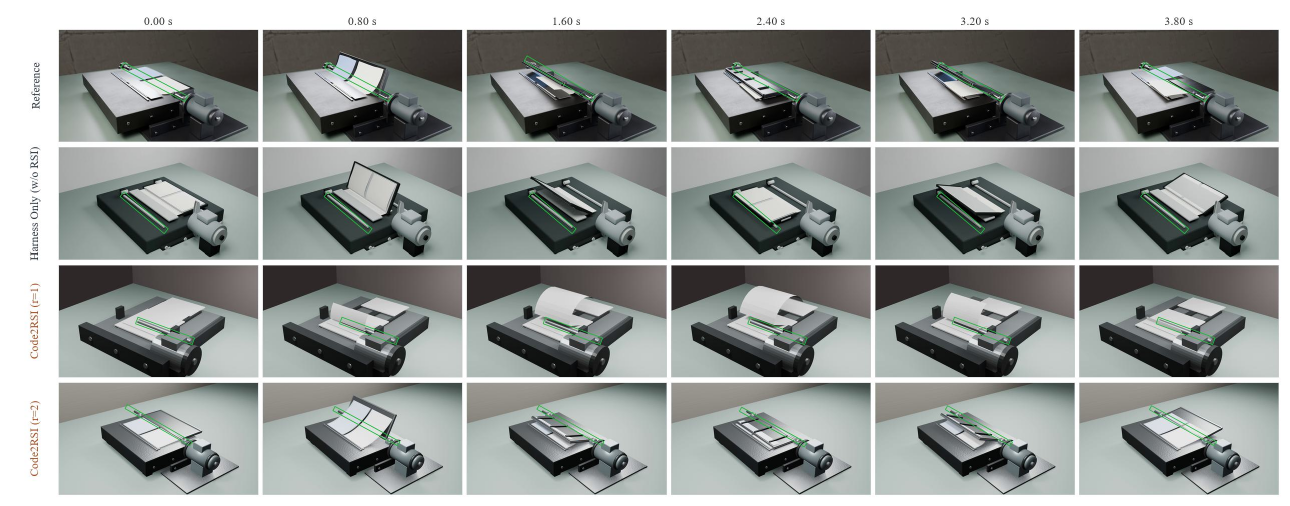}
  \caption{Dynamic Scene Reconstruction Across Two RSI Rounds. Green boxes highlight the central rod.}
  \label{fig:rsi-rod-keyframes}
\end{figure*}

\subsection{Results}

\subsubsection{The Gaming Domain}
\label{app:gaming}

The gaming world program is a set of game commands that builds a scene, supplies resources and specifies the task. A player agent then enters the world, acts toward the task goal and produces a trajectory of actions. Verification compares this trajectory with the completion condition of the task.

\textbf{Settings.} The game tasks are built in Minecraft. Each task includes a textual requirement and a set of construction requirements. The construction requirements cover the blocks of the scene, the supplied items, the starting inventory, and the products that must not be supplied at initialization. Based on the tasks from MineDojo~\cite{fan2022minedojo}, we build 12 test tasks in total, covering six types: navigate, chop, obtain, smelt, craft, and use.
Each type has one completion condition: navigation requires reaching the targets in order, and the other types require the target item to enter the inventory or to be placed in the world. 
For example, the craft task asks the player to craft one activator rail from six iron ingots, one redstone torch, and two sticks at a crafting table in a cave. A scene variant keeps the task and replaces the terrain blocks with those of the new scene. Navigation tasks additionally name the target block and require a distance of at least 20 blocks between the spawn point and the target. 

\begin{table}[t]
  \caption{Evaluation Results in the Gaming Domain. TC $\uparrow$ measures the proportion of tasks that the player completes in the constructed world. Static mean measures whether the program executes and contains the requested content.}
  \label{tab:gaming-eval}
  \centering
  \small
  \setlength{\tabcolsep}{6pt}
  \renewcommand{\arraystretch}{1.15}
  \begin{tabular}{@{}lcc@{}}
    \toprule
    Method & TC $\uparrow$ & Static mean \\
    \midrule
    Harness only & 0.417 & 82.8 \\
    Code2RSI (Round 1) & 0.667 & 100 \\
    Code2RSI (Round 2) & 0.750 & 100 \\
    \bottomrule
  \end{tabular}
\end{table}

The world program $P$ includes three files. The scene file places blocks. The task file supplies resources and the starting inventory. The metadata file records the instruction and the task type. 
The backend $E$ has two forms. 
During construction it is a Python interpreter inside the harness. The interpreter reads the two command files and executes the supported subset of Minecraft commands. It builds the block occupancy table, the player inventory and the container contents without launching the game. 
During gameplay $E$ is the official Minecraft 1.20.1 Java Edition server. The player is a Mineflayer bot that connects to this server and executes the actions chosen by LLMs. DeepSeek-V4.1-Flash is set as the default LLM for both world building and playing.

\subsubsection{The Embodied Manipulation Domain}
\label{app:embodied-results}

We also report a supplementary garment-manipulation study that is not covered in the main text, which evaluates task completion and physical consistency. The embodied world program controls arm motions and gripper states to carry out a garment manipulation instruction.
The harness executes the program and returns execution feedback to guide its revision.

\textbf{Settings.} We evaluate five garment manipulation tasks with DeepSeek-V4.1-Flash and GPT-6-Astra, comparing Harness (w/o RSI) with Code2RSI ($r{=}1$).
The world program $P$ is written in TypeScript.
The backend $E$ runs the generated motion in PyBullet, where two kinematic grippers manipulate a spring-based garment mesh.
Verification checks physical constraints, collisions, and grasp consistency, and reports violations and affected frames to the coding agent.

\textbf{Metrics.} We report final-state task completion (TC), constraint-violation frame rate (CV), and collision-check failure rate (Col.).
TC measures the proportion of executions whose final state satisfies the task completion condition.
CV is the fraction of frames that violate at least one physical constraint.
Col. is the fraction of collision checks in which penetration between a dynamic object and a static obstacle exceeds the collision tolerance.
We average each metric over episodes and report the equal-weight mean across the five tasks as a percentage.

Table~\ref{tab:physical-overview} shows that Code2RSI ($r{=}1$) lowers CV by 6.18 percentage points with DeepSeek-V4.1-Flash and 7.32 points with GPT-6-Astra.
Col. drops from 2.94\% to 0.35\% and from 0.34\% to 0\%, respectively, while TC remains at 100\% and 80\%.
The lower violation rates at unchanged task completion show that the retained harness improves the physical consistency of the generated motions.

\begin{table}[t!]
  \caption{Garment Manipulation Results with Harness (w/o RSI) and Code2RSI ($r{=}1$).
  Scores are equal-weight means across five garment tasks, expressed as percentages.
  TC: final-state task completion; CV: constraint-violation frame rate;
  Col.: collision-check failure rate.}
  \label{tab:physical-overview}
  \centering
  \small
  \setlength{\tabcolsep}{2pt}
  \renewcommand{\arraystretch}{1.18}
  \begin{tabular*}{\columnwidth}{@{\extracolsep{\fill}}llrrr@{}}
    \toprule
    Model & Method & TC $\uparrow$ & CV $\downarrow$ & Col. $\downarrow$ \\
    \midrule
    DeepSeek-V4.1-Flash & Harness (w/o RSI) & 100 & 45.1 & 2.94 \\
     & Code2RSI ($r{=}1$) & 100 & \textbf{38.92} & \textbf{0.35} \\
    \midrule
    GPT-6-Astra & Harness (w/o RSI) & 80 & 42.7 & 0.34 \\
     & Code2RSI ($r{=}1$) & 80 & \textbf{35.38} & \textbf{0} \\
    \bottomrule
  \end{tabular*}
\end{table}

\subsubsection{The Rendering Domain}
\label{app:rendering}

\begin{figure*}[t]
  \centering
  \includegraphics[width=\textwidth]{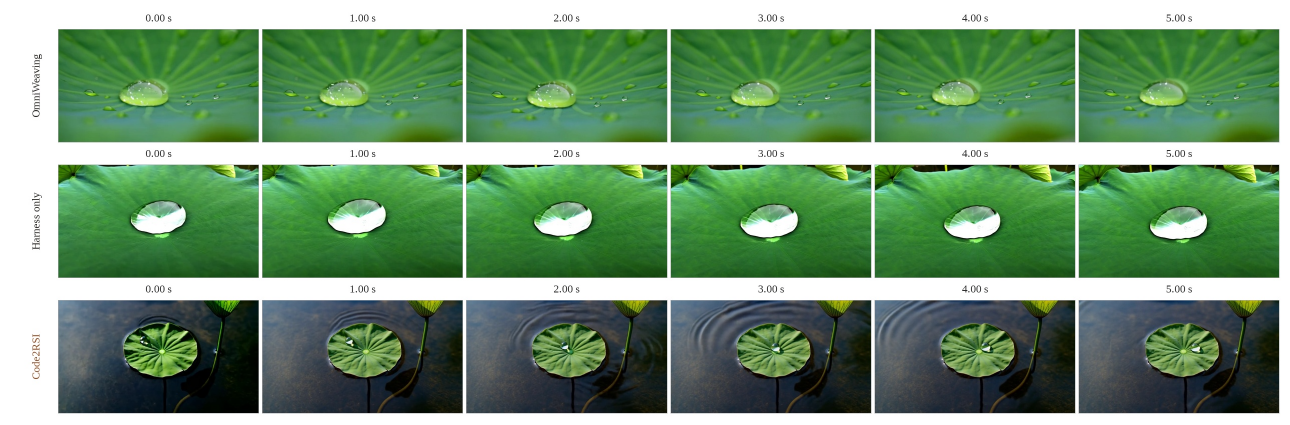}
  \caption{Visual Comparison of Code2RSI and Baselines on the Task of ``A glistening dewdrop is sliding gracefully across the textured surface of a lotus leaf, accentuating its shape as it moves.'' }
  \label{fig:phyg-dewdrop}
\end{figure*}

\begin{figure*}[t]
  \centering
  \includegraphics[width=\textwidth]{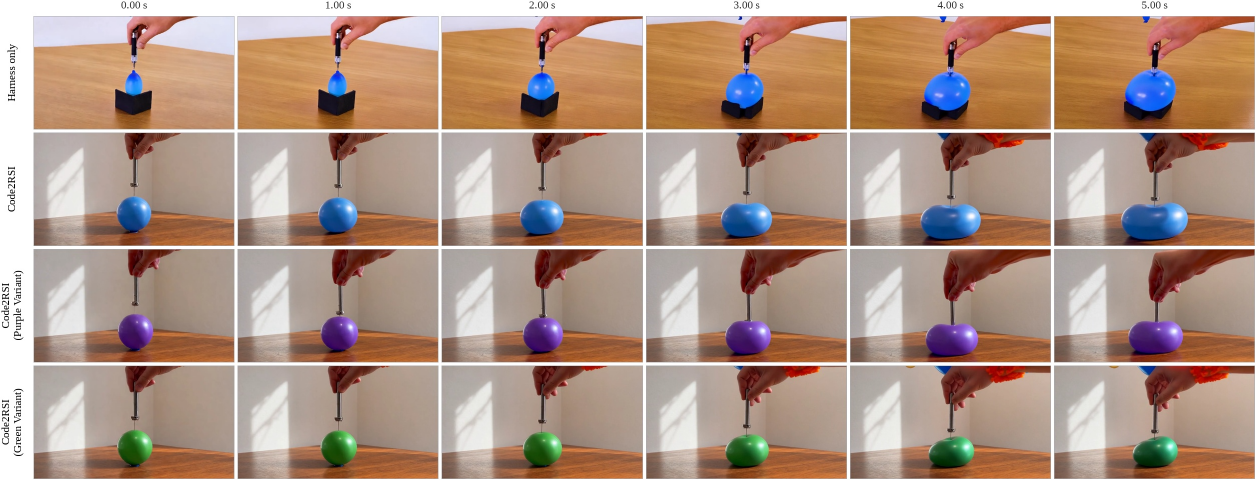}
  \caption{Visual Comparison of Code2RSI and Baselines on the Task of ``A hand slowly presses a needle into the surface of a fully inflated balloon, with the tip gradually penetrating the outer layer.'' The first two rows compare the fixed harness with Code2RSI. The last two rows demonstrate appearance control: the same gray-model video is used, while only the balloon colour in the refined reference frame is changed to generate videos of a needle puncturing balloons of different colours.}
  \label{fig:supp-phyg-balloon-control}
\end{figure*}

The rendering world program builds the scene geometry and its animation based on a Blender script.
The harness executes the script and records the evaluated geometry at sampled frames. Verification compares the recorded geometry with task-specific checks derived from the textual requirement. The constructed scene is then rendered as a gray-model video, which conditions a video generation model to produce the final video.

\textbf{Settings.} We extract 28 rendering tasks from PhyGenBench~\cite{meng2024towards}. The caption of each task is provided to the agent as the task description. 
The world program $P$ is the Python script that defines the scene construction function. The function builds low-poly gray geometry from primitives, places a camera and animates the objects. 
The backend $E$ has two stages. 
During construction $E$ is a headless Blender 4.4 process inside an isolated sandbox. The process clears the scene, calls the construction function and saves the scene file, which is rendered as a gray video after the final revision.
The second process reloads the saved file and records the observation $O$ at 9 frames from the first frame to frame 121 in steps of 15. 
The first frame of the rendered video is refined by the image model (GPT-image-2) into a colour reference. 
The gray video, the reference image and the caption are passed to the video generation model OmniWeaving~\cite{pan2026omniweaving} to produce the final video.

\textbf{Results.} Figures~\ref{fig:phyg-dewdrop} and~\ref{fig:supp-phyg-balloon-control} show qualitative comparisons on the dewdrop and balloon puncture tasks. Code2RSI generates scenes and motions that more closely follow the textual instructions than the baselines. Generating a gray-model video before appearance refinement also makes the final video more controllable by allowing appearance to be adjusted while keeping the underlying scene geometry and motion fixed. As shown in Figure~\ref{fig:supp-phyg-balloon-control}, reusing the same gray-model video and changing only the balloon colour in the refined reference frame produces videos of a needle puncturing differently coloured balloons while preserving the underlying puncture motion.

\end{document}